%% file: bml_arxiv.tex
\documentclass[final]{cvpr}
\usepackage{times}
\usepackage{epsfig}
\usepackage{graphicx}
\usepackage{amsmath}
\usepackage{amssymb}
\usepackage{amsthm}
\usepackage{mathrsfs}
\usepackage{bbding}
\usepackage{multirow}
\usepackage{color,soul}
\usepackage[table]{xcolor}
\usepackage{threeparttable}
\usepackage{subfigure}
\usepackage{enumitem}
\usepackage{appendix}
\usepackage[linesnumbered,lined,vlined,ruled,commentsnumbered]{algorithm2e}
\makeatletter
\newcommand{\thickhline}{
    \noalign {\ifnum 0=`}\fi \hrule height 1.2pt
    \futurelet \reserved@a \@xhline
}
\newcolumntype{x}[1]{>{\centering\let\newline\\\arraybackslash\hspace{0pt}}p{#1}}
\usepackage[pagebackref=true,breaklinks=true,colorlinks,bookmarks=false]{hyperref}

\def\cvprPaperID{1314} % *** Enter the ICCV Paper ID here
\def\confYear{ICCV 2021}

\begin{document}

%%%%%%%%% TITLE
\title{Binocular Mutual Learning for Improving Few-shot Classification}

\author{Ziqi Zhou$^{1}$ \quad
        Xi Qiu$^{1}$\thanks{\textit{Corresponding author}.}  \quad
        Jiangtao Xie$^{2}$  \quad
        Jianan Wu$^{1}$  \quad Chi Zhang$^{1}$\\
        $^{1}$Megvii Technology~~~$^{2}$Dalian University of Technology\\
        {\tt\small \{zhouziqi,qiuxi,wjn,zhangchi\}@megvii.com~~jiangtaoxie@mail.dlut.edu.cn}
}

\maketitle

%%%%%%%%% ABSTRACT
\begin{abstract}
Most of the few-shot learning methods learn to transfer knowledge from datasets with abundant labeled data (i.e., the base set). From the perspective of class space on base set, existing methods either focus on utilizing all classes under a global view by normal pretraining, or pay more attention to adopt an episodic manner to train meta-tasks within few classes in a local view. However, the interaction of the two views is rarely explored. As the two views capture complementary information, we naturally think of the compatibility of them for achieving further performance gains. Inspired by the mutual learning paradigm and binocular parallax, we propose a unified framework, namely \textbf{B}inocular \textbf{M}utual \textbf{L}earning (BML), which achieves the compatibility of the global view and the local view through both intra-view and cross-view modeling. Concretely, the global view learns in the whole class space to capture rich inter-class relationships. Meanwhile, the local view learns in the local class space within each episode, focusing on matching positive pairs correctly. In addition, cross-view mutual interaction further promotes the collaborative learning and the implicit exploration of useful knowledge from each other. During meta-test, binocular embeddings are aggregated together to support decision-making, which greatly improve the accuracy of classification. Extensive experiments conducted on multiple benchmarks including cross-domain validation confirm the effectiveness of our method\footnote{\url{https://github.com/ZZQzzq/BML}}.

\end{abstract}

%%%%%%%%% BODY TEXT
\input{Sec/Sec01/Introduction}
\input{Sec/Sec02/RelatedWork}

\input{Sec/Sec03/Method}
\input{Sec/Sec04/Experiments}

\section{Conclusion}
Inspired by mutual learning paradigm and binocular parallax, we propose a unified Binocular Mutual Learning (BML) framework, which achieves the compatibility of the global view and the local view through both intra-view and cross-view modeling. The effectiveness of BML has been fully demonstrated on both within-domain and cross-domain evaluations. The aggregated features are more robust than other competitors when dealing with degradation attacks. Besides, BML obtain accurate similarity ranking.

\noindent\textbf{Acknowledgement. } This work is supported by the National Key R\&D Plan of the Ministry of Science and Technology (Project No. 2020AAA0104400).

\begin{appendices}
\input{Sec/Supp/supp}
\end{appendices}

{\small
\bibliographystyle{ieee_fullname}
\bibliography{reference}
}

\end{document}

%% file: Sec/Sec01/Introduction.tex
\section{Introduction}\label{sec1}
Conventional classification methods heavily rely on massive labeled data~\cite{russakovsky2015imagenet} with diverse visual variations. However, in many realistic scenarios, only limited labeled data is available~\cite{yao2019learning,mahajan2020meta}, thereby giving rise to the investigation of few-shot classification (FSC), where only few available training data is given for the learning of new visual concepts. Such a setting makes FSC a challenging problem, since novel classes are unpredictable and the sampling of few shots is also biased. In order to overcome those difficulties, many effective approaches have been proposed in recent years, which can be mainly summarized into two categories according to training strategies. The first category is fine-tuning based paradigms~\cite{qiao2018few,lifchitz2019dense,chen2019closer,chen2020new,tian2020rethinking}, which learn classifiers in the whole base class space with a straightforward purpose of maximize differences between classes. Since all base classes are visible under each iteration, we refer to this kind of methods as the \emph{global} view. The other promising strategy is metric-based meta-training schemes~\cite{fei2006one,vinyals2016matching,snell2017prototypical,sung2018learning,oreshkin2018tadam,ye2020few,li2020boosting,guo2020attentive}, which only tune on a few classes in each episode. The main idea comes from metric learning and the purpose is to match unlabeled query to its correct class with a small labeled support set. Because the visible range of base classes for each meta-task is limited, we oppositely call them the \emph{local} view.

\begin{table}[t]
    \centering
    \caption{Comparison of BML with several representative methods from the view of base class space. Accuracy (Acc.) from \emph{mini}ImageNet~\cite{vinyals2016matching}. Obviously, the unified perspective of BML is more effective.}
    \vspace{2pt}
    \label{Tab:BML_compare}
    \small
    \begin{tabular}{p{65pt}|p{25pt}p{25pt}p{36pt}p{17pt}}
    \thickhline
    &\emph{w/}global&\emph{w/}local&\emph{strategy}&\emph{Acc.}
    \\ \thickhline
    MAML~\cite{finn2017model}&\multirow{5}{*}{\XSolidBrush}&\multirow{5}{*}{\CheckmarkBold}&\emph{one-stage}&63.1
    \\
    MatchingNet~\cite{vinyals2016matching}&&&\emph{one-stage}&55.3
    \\
    RelationNet~\cite{sung2018learning}&&&\emph{one-stage}&65.3
    \\
    ProtoNet~\cite{snell2017prototypical}&&&\emph{one-stage}&68.2
    \\ \thickhline
    Rethink~\cite{tian2020rethinking}&\multirow{3}{*}{\CheckmarkBold}&\multirow{3}{*}{\XSolidBrush}&\emph{one-stage}&82.1
    \\
    DC~\cite{lifchitz2019dense}&&&\emph{one-stage}&79.0
    \\
    CloserLook~\cite{chen2019closer}&&&\emph{one-stage}&75.7
    \\ \thickhline
    DeepEMD~\cite{zhang2020deepemd}&\multirow{4}{*}{\CheckmarkBold}&\multirow{4}{*}{\CheckmarkBold}&\emph{two-stage}&82.4
    \\
    FEAT~\cite{ye2020few}&&&\emph{two-stage}&82.1
    \\
    Meta-Baseline~\cite{chen2020new}&&&\emph{two-stage}&79.3
    \\
    Neg-Cosine~\cite{liu2020negative}&&&\emph{two-stage}&81.6
    \\ \thickhline \rowcolor{gray!20}
    \textbf{Our BML}&\CheckmarkBold&\CheckmarkBold&\emph{\textbf{one-stage}}&\textbf{83.6}
    \\ \thickhline
    \end{tabular}
\end{table}
Considering that single view (whether \emph{global} or \emph{local}) is relatively weak, it is not enough to provide adequate knowledge for accurate classification. What's more, the combination of dual views fits well with the characteristics of ``people deepen their perception through two eyes". To this end, we propose this new \textbf{B}inocular \textbf{M}utual \textbf{L}earning (BML) paradigm, which equips the network with a global view and a local view simultaneously. The  combination of two complementary views works like a binocular system, and the mutual interaction~\cite{zhang2018deep} through two views further promotes their cooperation and calibrates the inappropriate expression caused by single ``biased" view. Concretely, BML generates better expression through both intra-view and cross-view modeling. The intra-view training captures view-specific knowledge, where two balanced feature space are built, one focuses on inter-class relationship perception (global view) and the other pays attention on matching details (local view). Meanwhile, the cross-view mutual interaction facilitates the implicit knowledge transfer from each other. To balance ``binocular parallax", we enlarge the optimization difficulty of the local view, so that the global view can learn more useful knowledge from mutual interaction (For more details, please refer to Section.\ref{sec3.2}).

Clearly, BML paradigm has two advantages: strong transferability and high time-efficiency, which are manifested from the following comparisons. Concretely, compared with single view based methods mentioned above, BML has two complementary views, so that more transferable and reliable knowledge can be learned. In contrast, single global training lacks additional constraints, making it easy to over-fit on base patterns~\cite{goldblum2020unraveling}. Meanwhile, single local training is restricted by local perspectives, whose performance is heavily depend on the configuration of tasks, not to mention the complex structures. Moreover, compared with two-stage methods which firstly excute global training and then tune the embedding with local training, BML unifies the two views in a one-stage framework and enables the promotion of each other. By contrast, two-stage methods~\cite{sun2019meta,chen2019closer,zhang2020deepemd,ye2020few,liu2020negative} are time-consuming. They focus on how to better learn embedding in the first stage to provide stronger features for the second stage of optimization~\cite{liu2020negative}, but ignore that local view and global view can promote each other in a unified manner.

We highlight the advantages of BML compared with several typical approaches in Table~\ref{Tab:BML_compare}. As the first batch of methods to consider the combination of dual views, we propose an elegant compatibility strategy: binocular mutual learning, which is inspired by the fact that human beings usually perceive the world through two eyes (benefit from appropriate binocular parallax). The two complementary views simulate the binocular mode, and the mutual interaction calibrates deviation. Extensive experiments confirm the effectiveness of BML, which is mainly reflected on the stable performance under different granularity evaluation. No matter facing coarse-grained (e.g., \emph{mini}ImageNet~\cite{vinyals2016matching}) or fine-grained (e.g. CUB~\cite{WahCUB_200_2011}) situation, BML performs well. However, single view based methods cannot handle all the situations. This confirm that the unified framework does facilitate the mutual calibration of the two views. In summary, the contributions of this paper are as follows:
\begin{itemize}[itemsep=-5pt,topsep=5pt]
    \item We closely analyze the status quo of FSC and propose an efficient one-stage \textbf{B}inocular \textbf{M}utual \textbf{L}earning paradigm: \textbf{BML}, which elegantly aggregate the global view and the local view through both intra-view and cross-view modeling.
    \item To enhance mutual learning, we propose an elastic loss to readjust the optimization difficulty of the local view, which promotes the bidirectional implicit knowledge transfer.
    \item Extensive experiments on multiple benchmarks including cross-domain validation verify the effectiveness of our framework.
\end{itemize}

%%%%%%%%% BODY TEXT

%% file: Sec/Sec02/RelatedWork.tex
\section{Related Works}
\subsection{Fine-Tuning based Methods (\emph{global} view)}
Researches represented by~\cite{qiao2018few,sun2019meta,lifchitz2019dense,chen2019closer,tian2020rethinking,liu2020negative} pay more attention to global training by simply learning a class-specific embedding with fully-connected (FC) layers. Among them, FSIR \cite{qiao2018few} adapted the pretrained model to the new categories by directly predicting the parameters from activation, while DC~\cite{lifchitz2019dense} started from the perspective of spatial information mining and performed dense classification. MTL~\cite{sun2019meta} further enhanced the pretrained model through a hard task meta-batch mechanism. And~\cite{chen2019closer} proposed two different classification layers, including a conventional FC Layer (CloserLook) and a FC layer with feature normalization (CloserLook++). To alleviate over-fitting on base patterns, The authors in \cite{tian2020rethinking} employed self-distillation strategy and data augmentation to constraint the learning process.

\begin{figure*}[ht]
   \centering
   \includegraphics[width=\linewidth]{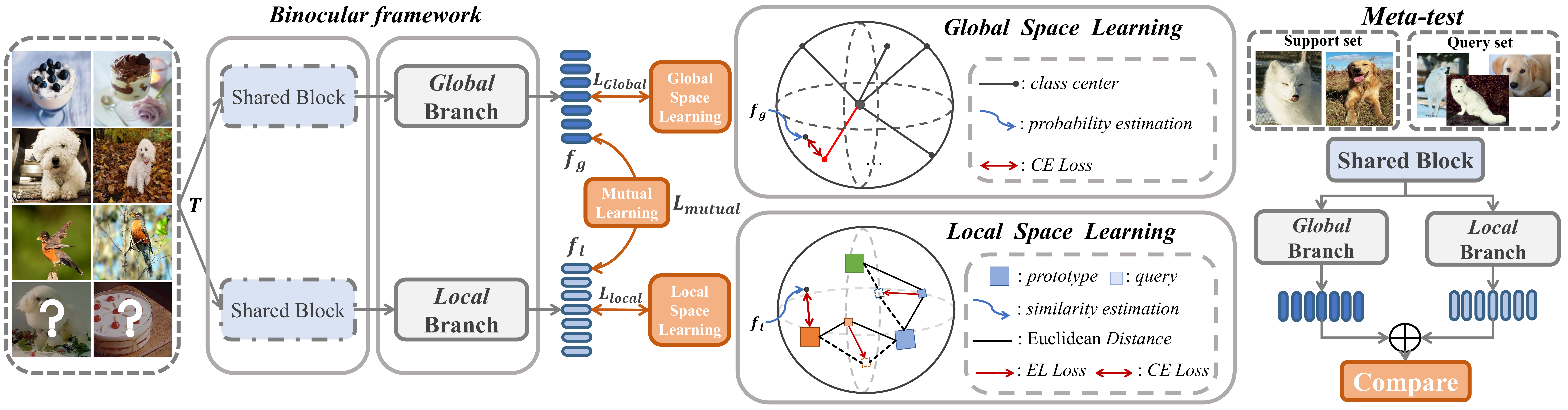}
   \caption{The framework of \textbf{BML}. For each task $\mathcal{T}$, BML minimizes the global classification loss $\mathcal{L}_{global}$, the local matching loss $\mathcal{L}_{local}$ and the distribution consistency loss $\mathcal{L}_{mutual}$ to capture discriminative expression. During test, the collaborative features from two views are combined to support the final decision.}
\label{fig02_framework}
\end{figure*}

\subsection{Meta-Training based Methods (\emph{local} view)}
One of the promising branch of meta-training is metric-based methods~\cite{vinyals2016matching,snell2017prototypical,sung2018learning,ye2020few,oreshkin2018tadam,li2020boosting,li2020adversarial,li2019finding,yang2020dpgn,hou2019cross,guo2020attentive,boudiaf2020transductive,qiao2019transductive}, which meta-learn an ideal metric space $\mathcal{S}(., .)$ to bring homogeneous samples closer while push heterogeneous samples away. The classification process is carried out under the guidance of the ``\emph{Nearest Neighbor}" principle. Specifically, Matching Networks~\cite{vinyals2016matching} proposed a LSTM~\cite{hochreiter1997long}-based encoding module to re-code the context of support feature, and computed attention scores using cosine distance. Similar variants include FEAT~\cite{ye2020few}, which replaced LSTM with transformer~\cite{vaswani2017attention}. Prototypical Networks~\cite{snell2017prototypical} employed prototype to identify each class and calculated the Euclidean distance. Relation Networks~\cite{sung2018learning} further adopted a learnable correlation calculation module to measure pairwise similarity. Subsequent follow-uppers derived many variants with incorporating cross-modal information~\cite{li2020boosting}, introducing adversarial noise~\cite{li2020adversarial}, employing local descriptors~\cite{li2019revisiting} or mutual information~\cite{guo2020attentive,boudiaf2020transductive}, making pretext tasks~\cite{zhang2021iept}, using attention mechanisms~\cite{fei2021melr,hou2019cross} or learning task-relevant metrics~\cite{oreshkin2018tadam,li2019finding}.

Besides, some studies started from optimization~\cite{finn2017model,li2017meta,jamal2019task,raghu2019rapid,lee2019meta}. A typical literature is MAML~\cite{finn2017model}, in which, the parameters of a base learner were further optimized in a few iterations for quickly adaption for new tasks. Subsequent variants further developed MAML by designing more objectives~\cite{li2017meta}, employing better classifiers~\cite{lee2019meta} or dynamically adjusting the weights of different tasks~\cite{jamal2019task}. More interestingly, the working mechanism of MAML was analyzed in detail in~\cite{raghu2019rapid} and the experiments highlighted that ``\emph{Feature Reuse}" is the key to its role. In other words, the performance is majorly determined by the quality of the learned features, which is revalidated by experiments conducted in this paper.

We also found that some recent researches~\cite{liu2020negative,chen2020new,zhang2020deepemd,ye2020few} have realized the promotion role of two views. Some of them~\cite{liu2020negative,chen2020new,zhang2020deepemd,ye2020few} employed a two-stage training scheme. They firstly pretrained under global mode, and then tuned the parameters using local mode. Others~\cite{chen2020diversity,hou2019cross,oreshkin2018tadam} promoted local training by learning additional global classifiers. All of them focus on the promotion of global to local but ignore the bidirectional cross-view interaction (which plays an important role in BML).

\subsection{Mutual Learning}
Mutual learning~\cite{zhang2018deep} is a new distillation modal that has shined in many fields recently, which breaks the conventional ``teacher-student" structure which has fixed direction of supervision. In mutual learning, a group of students are collaboratively learned from each other, which helps to obtain more general models without pretrained teachers. Similar ideas have been employed in person re-ID~\cite{ge2020mutual}, but not effectively applied in FSC. Literatures like~\cite{tian2020rethinking,li2020few} are close to this topic, but the training process is $N$-staged with fixed pretrained teachers.

Inspired by mutual learning~\cite{zhang2018deep} and binocular parallax, we propose this unified BML framework with two complementary views. Each of them can be regarded as a ``\emph{student}". During training, besides their offline hard supervision, two ``\emph{students}" also learn collaboratively and implicitly explore useful knowledge from each other.

%% file: Sec/Sec03/Method.tex
\section{Methodology}~\label{sec3}
\subsection{Preliminary}
In the standard FSC scenario, we are given a base set with $\mathcal{C}_{base}$ classes and a novel set with $\mathcal{C}_{novel}$ classes, where $\mathcal{C}_{base}\cap \mathcal{C}_{novel} = \varnothing$. Training is usually performed on $\mathcal{C}_{base}$ classes and the optimization goal is to transfer the learned knowledge to new tasks built on $\mathcal{C}_{novel}$. During meta-test, a family of tasks $\{\mathcal{T}\}_1^n$ are constructed for evaluation. Concretely, each task $\mathcal{T}$ has a support set $\mathbb{S}$ and a query set $\mathbb{Q}$. The support set $\mathbb{S}$ contains $N$ classes and each class has $K$ images (\emph{i.e.}, the $N$-way $K$-shot setting). The query set $\mathbb{Q}$ includes $N\times Q$ unlabeled images. In most literature, $N$ is set to 5 and $K$ is set to 1 or 5, so do we.

\subsection{Binocular Mutual Learning}
As shown in Figure~\ref{fig02_framework}, BML has two complementary branches (views) based on shared blocks. In which, the \emph{global} branch learns in the whole class space for inter-class relationship mining, and the \emph{local} branch learns in each episode within few classes, aiming at matching each query sample to its support prototype. Besides, the two branches implicitly explore useful knowledge from each other by minimizing KL-Divergence based mimicry loss to match the feature distribution of its peers.

\subsubsection{Global Intra-view Training}
For the \emph{global} branch, the learned features are related to the whole class space, which explicitly contain rich inter-class relationships.

Specifically, given task $\mathcal{T}$, we learn a global learner $\mathcal{A}_\phi^G$ to map each image $\mathbf{x}_i$ in $\mathcal{T}$ to a high-dimensional space, and then a $1\times 1\times \mathcal{C}_{base}$ convolution layer $\mathcal{W}$ is learned to classify each point of the feature to its corresponding class. Let $w$ denotes the width of the feature and $h$ denotes the height, the probability estimate of point $(p,q)$ is formulated by:
\begin{equation}
P(y_i=y|\mathbf{x}_i^{(p,q)})= \sigma (\mathcal{W}(\mathcal{A}_\phi^G(\textbf{x}_i^{(p,q)})))
\label{Eq1}
\end{equation}
where $\sigma: \mathbb{R}^m \rightarrow \mathbb{R}^m$ is the \emph{softmax function}, $m$ represents the dimension size which is 640 here.

The negative logarithm of $P(y_i=y|\mathbf{x}^i_{(p,q)})$ is calculated to represent the loss of current feature point $(p,q)$ of input data $x_i$. The total global loss is the average value of all images in $\mathcal{T}$. Formulaically,
\begin{equation}
\mathcal{L}_{global} = \underset{(x_i, y_i) \in \mathcal{T}}{\mathbb{E}} \frac{1}{w\times h} \sum_{\substack{p=1\\q=1}}^{w,h} - y_i \mathbf{log} P(y_i=y|\mathbf{x}_i^{(p,q)})
\label{Eq2}
\end{equation}

Meaningfully, classifying each feature point correctly is helpful to capture spatial structure information of the foreground. Since each point can be traced back to a local area of the foreground, multiple points represents multiple local areas that is equivalent to a memory-saving multi-crop operation~\cite{lifchitz2019dense}. For fair comparison, we also apply point-wise classification to the baselines we compared in this paper.

\subsubsection{Local Intra-view Training}\label{sec3.2}
The \emph{local} branch borrows the idea of metric learning and learns to match each query sample with support prototypes in embedding space.

For task $\mathcal{T}$ with $N$ classes, we divide it into a support set $\mathbb{S}=\{(x_j, y_j)\mid j=1,\cdots,N\times S)\}$ and a query set $\mathbb{Q}=\{(x_j, y_j)\mid j=1,\cdots,N\times Q)\}$. A local learner $\mathcal{A}_\phi^L$ is learned to map all samples $\{x_j\mid x_j \in \mathbb{S}\cup\mathbb{Q}\}$ to an embedding space, and the matching process is guided by nearest neighbor strategy, where $N$ prototypes are calculated, \emph{i.e.}, $C_i=\frac{1}{S}\sum_{j\in\mathbb{S}_i}\mathcal{A}_\phi^L(x_j)$. To metric the similarity between query and each prototype, we simply employ Euclidean distance indicated by $\mathcal{M}$. The negative logarithm of all query samples are calculated to get the local loss.
\begin{equation}
P(C_i=C|\mathbf{x}_i)= \frac{\mathbf{exp} [-\mathcal{M}(C_i, \mathcal{A}_\phi^L(\textbf{x}_i))]}{\sum_{j=1}^N \mathbf{exp}[-\mathcal{M}(C_j, \mathcal{A}_\phi^L(\textbf{x}_i))]}
\label{Eq3}
\end{equation}

\begin{equation}
\mathcal{L}_{local} = \underset{(x_i, y_i) \in \mathbb{Q}}{\mathbb{E}} - C_i \mathbf{log} P(C_i=C|\mathbf{x}_i)
\label{Eq4}
\end{equation}

\textbf{Elastic constraint for magnifying optimization difficulty.} Moreover, we consider the following two points and further optimize the local loss by applying an elastic constraint.

On the one hand, each task $\mathcal{T}$ is a collection of randomly sampled data, resulting in randomness of difficulty. Treating all tasks equally during cross-epoch training like~\cite{snell2017prototypical} will lead to an ``unhealthy" phenomenon: further performance improvement is hard to obtain in the later training, because the model is dynamically growing while the difficulty is static~\cite{bengio2009curriculum}. On the other hand, the optimization difficulty of the local branch is relatively simple compared with the global one (since each query only has $N-1$ negative prototypes while global branch has $\mathcal{C}_{base}-1$). The implementation of mutual promotion requires a seedbed, that is, the two views can provide valuable knowledge to each other.
\begin{figure}[ht]
   \centering
   \includegraphics[width=\linewidth]{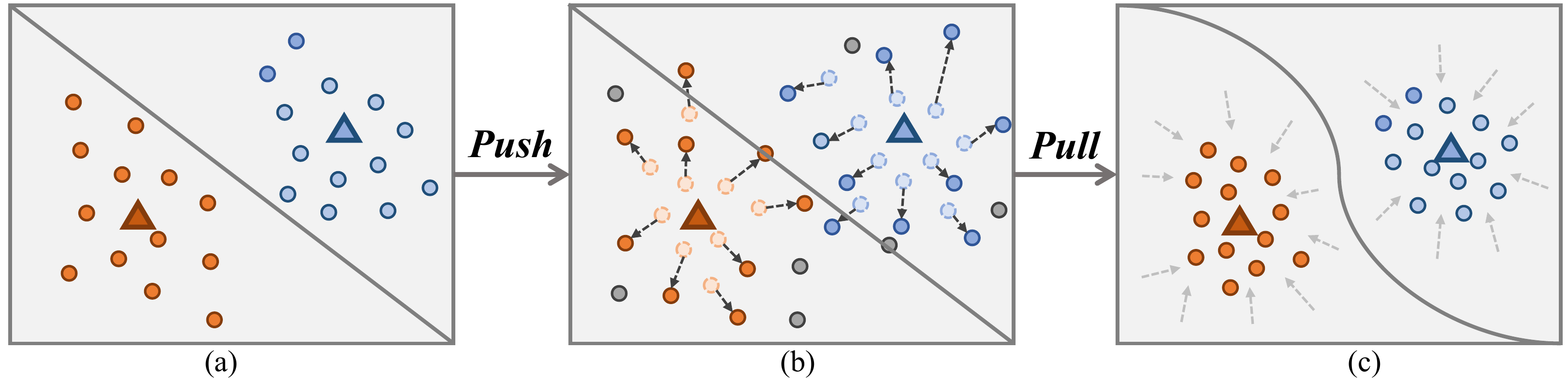}
   \caption{The process of elastic loss. All queries are pushed away from their clusters to magnify the optimization difficulty, and the network is forced to pull them back again.}
\label{fig03}
\end{figure}

Therefore, we propose this elastic loss to enhance the training difficulty of the local branch. The complete process is shown in Figure~\ref{fig03}. Simply, we modify the position of each query sample $\mathbf{x}_i$ in the embedding space according to the difference between $\mathcal{M}(C_p, \mathcal{A}_\phi^L(\mathbf{x}_i))$ and $\mathcal{M}(C_n, \mathcal{A}_\phi^L(\mathbf{x}_i))$, where $C_p$ is the positive prototype and $C_n$ is the nearest negative prototype. The above operation can be described as ``\textbf{\emph{push away}}". Then we demand the network to \textbf{\emph{pull}} these pushed out samples \textbf{\emph{back}} to the vicinity of their positive prototypes, thus to learn more implicit knowledge and get further performance improvement.

To better perceive the foreground from different patches and learn rich relationship between $C_p$ and patch-specific $C_n^{(p,q)}$, similar to the global branch, instead of performing global average pooling, we calculate elastic constraint from different spatial position. Forcing the network to pull $\mathbf{x}_i$ to its positive prototype under different ($C_n^{(p,q)}$, $C_p$) pairs helps to mine more hard samples, and avoid over-fitting on simple local tasks.

The updated local loss has a formulation as:
\begin{equation}
\hat{P}(C_i=C|\mathbf{x}_i^{(p,q)})= \frac{\mathbf{exp} -\mathcal{M}(C_i, \mathcal{A}_\phi^L(\textbf{x}_i^{(p,q)}))-d_{\text{EL}}^{i_{(p,q)}}}{\sum_{j=1}^N \mathbf{exp}-\mathcal{M}(C_j, \mathcal{A}_\phi^L(\textbf{x}_i^{(p,q)}))}
\label{Eq5}
\end{equation}
\begin{equation}
\hat{\mathcal{L}}_{local} = \underset{(x_i, y_i) \in \mathbb{Q}}{\mathbb{E}} \frac{1}{w\times h}\sum_{\substack{p=1\\q=1}}^{w, h}- C_i \mathbf{log}\hat{P}(C_i=C|\mathbf{x}_i^{(p,q)})
\label{Eq6}
\end{equation}
where $d_{\text{EL}}^{i_{(p,q)}}$ is the patch-specific elastic constraint, whose specific detail is described in Algorithm~\ref{alg02}. $\alpha_1$ and $\alpha_2$ are two scale factors, $\alpha_1$ adjusts the push degree cross-epochs and $\alpha_2$ adjusts the push degree cross-tasks.
\input{Sec/Sec03/3_1_algorithm}

\subsubsection{Cross-view Mutual Learning}
In addition to separate intra-view learning, the two views also promote each other through cross-view mutual interaction. Specifically, for each view, in addition to completing its own offline hard tasks, it is also forced to minimize the mimicry loss from another view (based on KL-divergence), which encourage the implicit knowledge transfer. Clearly, the mutual loss has two sub-items, which are formulated as:
\begin{equation}
\mathcal{L}_{mutual} = \mathcal{D}_{\text{KL}}(\text{F}_l||\text{F}_g) + \mathcal{D}_{\text{KL}}(\text{F}_g||\text{F}_l)
\label{Eq7}
\end{equation}

\begin{equation}
\mathcal{D}_{\text{KL}}(\text{F}_g||\text{F}_l)=\text{F}_g\text{log}\frac{\text{F}_g}{\text{F}_l}, \mathcal{D}_{\text{KL}}(\text{F}_l||\text{F}_g)=\text{F}_l\text{log}\frac{\text{F}_l}{\text{F}_g}
\label{Eq8}
\end{equation}
where $\text{F}(\cdot)$ represents the feature distribution calculated by $\sigma(\mathcal{A}_\phi(x))$. We consider the issue of interaction from the perspective of feature distribution consistency, and only learn relative relationships rather than hard constraints such as mean squared error in Euclidean space. This is because that too strong supervision signal is harmful to retain the specificity of the two views.

In summary, the final loss has following formulation:
\begin{equation}
\mathcal{L}_{total} = \alpha\mathcal{L}_{global} + \beta \mathcal{L}_{local} + \gamma \mathcal{L}_{mutual}
\label{Eq9}
\end{equation}
where $\alpha$, $\beta$ and $\gamma$ are weighting factors.

\subsubsection{Inference}
During meta-test, we integrate the results of the global and local branches. We do not perform global average pooling but directly flatten the features and calculate the Euclidean distance (the same as ProtoNet~\cite{snell2017prototypical}). The integration can be done on the feature level or on the logits level. In this paper, we simply integrate the global and local logits.

%% file: Sec/Sec03/3_1_algorithm.tex
\begin{algorithm}[h]
\caption{Elastic Constraint $d_{EL}^{(p,q)}$ on $x_i^{(p,q)}$.}
\SetAlgoLined
\label{alg02}
\KwIn{prototypes $\{C_i^{(p,q)}\}_i^N$; query $\mathcal{A}_\phi^l(x_i^{(p,q)})$; current epoch $e$; total epoch $E$; $\alpha_1, \alpha_2$}
\KwOut{Elastic Constraint $d_{EL}^{(p,q)}$}
\BlankLine
Get logits $L^{(p,q)}$ of $\mathcal{A}_\phi^l(x_i^{(p,q)})$ using Eq.\ref{Eq3};\\
Cal distance between $\mathcal{A}_\phi^l(x_i^{(p,q)})$ and $C_p$: {\small $\emph{dis}_P=\mathbb{I}_{c=C_p}L^{(p,q)}$};\\
Cal distance between $\mathcal{A}_\phi^l(x_i^{(p,q)})$ and $C_n^{(p,q)}$: {\small $\emph{dis}_N=\textbf{sort}(\mathbb{I}_{c\neq C_p}L^{(p,q)})[0]$};\\
Cal $\Delta=\emph{dis}_P-\emph{dis}_N$;\\
Return $d_{EL}^{(p,q)} = \frac{\alpha_1 \cdot e/E}{1+\mathbf{exp}(-\alpha_2 \cdot\Delta)}$;\\
\end{algorithm}

%% file: Sec/Sec04/Experiments.tex
\section{Experiments}
In this section, we answer the following questions:
\begin{itemize}[itemsep=-5pt,topsep=-2pt]
    \item How does our BML perform compared with SoTAs?
    \item Why is binocular learning better?
    \item How does the elastic loss work?
    \item Is BML less sensitive to image quality?
    \item Should the standards for validation be diverse?
\end{itemize}

\subsection{Meta Datasets}
We validate our BML on four commonly used benchmarks, including \emph{mini}ImageNet~\cite{vinyals2016matching}, \emph{tiered}ImageNet~\cite{ren2018meta}, CIFAR-FS~\cite{bertinetto2018meta} and CUB-200-2011 (CUB)~\cite{WahCUB_200_2011}. Details of those datasets are summerized in Table~\ref{Tab:dataset}. All the input images are resized to $84\times84$ during comparison, and in the ablation part, we further analyze the stability of BML when enlarging the image size to $224\times224$ or introducing other degraded components. In particular, for CUB, we crop the test images with the bounding box provided by~\cite{triantafillou2017few} to make a fair comparison.
\begin{table}[h]
    \centering
    \caption{Summarization of four benchmarks. }
    \vspace{2pt}
    \label{Tab:dataset}
    \small
    \begin{tabular}{c|c|c|c}
    \thickhline
    &Images&Classes&Split
    \\ \thickhline
    \emph{mini}ImageNet~\cite{vinyals2016matching}&60,000&100&64/16/20
    \\
    \emph{tiered}ImageNet~\cite{ren2018meta}&779,165&608&351/97/160
    \\
    CIFAR-FS~\cite{bertinetto2018meta}&60,000&100&60/16/20
    \\
    CUB-200-2011~\cite{WahCUB_200_2011}&11,788&200&100/50/50
    \\ \thickhline
    \end{tabular}

\end{table}

\subsection{Implementation Details}
\textbf{Architecture.} Following previous works~\cite{oreshkin2018tadam,tian2020rethinking,lee2019meta}, we use ResNet12 as our backbone, which consists of 4 residual blocks. Each block has 3 convolutional layers with 3$\times$3 kernel and a 2$\times$2 max-pooling layer. We remove the last global average pooling layer to preserve spatial information. Similar to~\cite{lee2019meta}, we use Dropblock as a regularizer and the number of filters are set to (64, 160, 320, 640). Since BML have a binocular structure, we share the first three blocks and assign an independent block-4 to each \emph{view}.
\input{Sec/Sec04/4_1_imagenet_serious}

\textbf{Optimization setup.} We use SGD optimizer with a momentum of 0.9 and a weight decay of 5e$-$4. The learning rate is initialized as 0.1. For \emph{mini}ImageNet, CIFAR-FS and CUB, we train 100 epochs. In the 50$-th$ epoch, the learning rate is reduced to 6e$-$3, and further reduced to 1.2e$-$4 in the 70$-th$ epoch. For \emph{tiered}ImageNet, we train 150 epochs and decay the learning rate by 0.1 times per 40 epochs. In particular, when organizing data, in order to adapt to \emph{binocular} requirements, we adopt a uniform sampling strategy. The ratio in the loss is set to $\alpha$:$\beta$:$\gamma$=4:2:1, and the two scale factors $\alpha_1$ and $\alpha_2$ in the elastic loss are experimentally set to 5.5 and 0.1, respectively. To ensure the stability of the evaluation results, for each benchmark, we test 2,000 episodes and report the average performance.

\subsection{Experimental Results}

\subsubsection{Comparison on Coarse-grained Benchmark}\label{sec4_1_1}
Comparison to prior works are shown in Table~\ref{Tab:ImageNetserious}, our results are highlighted in bold with gray background. Specific structural details of \emph{ConvNet} are reported as follows (filter number of four blocks): MAML~\cite{finn2017model}: 32-32-32-32; TAML~\cite{jamal2019task}, ProtoNet~\cite{snell2017prototypical}, MatchingNet~\cite{vinyals2016matching}: 64-64-64-64; RelationNet~\cite{sung2018learning}: 64-96-128-256.

As is reported, compared with two baselines, the performance of BML is remarkable, even 9\% higher in some cases (details is analyzed in Section.\ref{sec4_2}). Compared with other competitors, BML achieves a new SoTA on \emph{mini}ImageNet, including the best metric-based method FEAT~\cite{ye2020few}. On \emph{tiered}ImageNet, we also achieve good performance by simply using the \emph{nearest neighbor} principle. As for CIFAR-FS, we surpass all competitors and reach a new SoTA, including LR+ICI~\cite{wang2020instance} which is based on the transductive strategy.

\subsubsection{Comparison on Fine-grained Benchmark}\label{sec4_1_2}
\textbf{Within domain evaluation} results is shown in Table~\ref{Tab:CUB} (Results with $^\dagger$ are reported in \cite{chen2019closer}). BML outperforms the runner-up DeepEMD~\cite{zhang2020deepemd} by 1.76\% and 0.56\% at 5-shot and 1-shot setting respectively. Although DeepEMD~\cite{zhang2020deepemd} adopts the similar idea of spatial information mining, but the task-dependent patch-wise matching is time-luxurious, while our BML shifts the attention to the bottom embedding, which is time-efficient and valid.
\begin{table}[h]
    \centering
    \caption{Within domain comparison on CUB-200-2011.}
    \vspace{2pt}
    \label{Tab:CUB}
    \small
    \scalebox{0.9}{
    \begin{tabular}{l|c|c|c}
    \thickhline
    \multirow{2}{*}{Method}&\multirow{2}{*}{Backbone}&\multicolumn{2}{c}{CUB-200-2011}
    \\ \cline{3-4}
    &&\emph{5-way 1-shot}&\emph{5-way 5-shot}
    \\ \thickhline
    MAML$^\dagger$~\cite{finn2017model}&\emph{ResNet18}&68.42$\pm$1.07&83.47$\pm$0.62
    \\ \thickhline
    ProtoNet$^\dagger$~\cite{snell2017prototypical}&\emph{ResNet18}&72.99$\pm$0.88&86.64$\pm$0.51
    \\
    MatchingNet$^\dagger$~\cite{vinyals2016matching}&\emph{ResNet18}&73.49$\pm$0.89&84.45$\pm$0.58
    \\
    RelationNet$^\dagger$~\cite{sung2018learning}&\emph{ResNet18}&68.58$\pm$0.94&84.05$\pm$0.56
    \\
    DeepEMD~\cite{zhang2020deepemd}&\emph{ResNet12}&75.65$\pm$0.83&88.69$\pm$0.50
    \\ \thickhline
    CloserLook~\cite{chen2019closer}&\emph{ResNet18}&47.12$\pm$0.74&64.16$\pm$0.71
    \\
    CloserLook++~\cite{chen2019closer}&\emph{ResNet18}&60.53$\pm$0.83&79.34$\pm$0.61
    \\
    Centroid~\cite{afrasiyabi2020associative}&\emph{ResNet18}&74.22$\pm$1.09&88.65$\pm$0.55
    \\ \thickhline
    \textbf{Baseline-\emph{local}}&\emph{ResNet12}&66.79$\pm$0.49&86.55$\pm$0.28
    \\
    \textbf{Baseline-\emph{global}}&\emph{ResNet12}&60.13$\pm$0.49&79.77$\pm$0.36
    \\ \rowcolor{gray!20}
    \textbf{BML}&\emph{ResNet12}&\textbf{76.21$\pm$0.63}&\textbf{90.45}$\pm$\textbf{0.36}
    \\ \thickhline
    \end{tabular}
    }
\end{table}

\input{Sec/Sec04/4_0_binocular_ablation}
What's more, comparing the performance of baseline-\emph{global} reported in Tables~\ref{Tab:ImageNetserious}-\ref{Tab:CUB} and Figure~\ref{Fig:shots_ablation}, we find that the performance of baseline-\emph{global} is significantly reduced on fine-grained benchmark CUB, which shows that single global training loses its effect when the inter-class difference is relatively small. However, our BML performs well on all granularities, which proves that binocular framework is more robust against granularity change.

\begin{table}[h]
    \centering
    \caption{Cross-domain comparison on CUB-200-2011. }
    \vspace{2pt}
    \label{Tab:mini2CUB}
    \small
    \begin{tabular}{l|c}
    \thickhline
    Method&\emph{mini}ImageNet $\to$ CUB
    \\ \thickhline
    MatchingNet~\cite{vinyals2016matching}&53.07$\pm$0.74
    \\
    ProtoNet$^\dagger$~\cite{snell2017prototypical}&62.02$\pm$0.70
    \\
    MAML$^\dagger$~\cite{finn2017model}&51.34$\pm$0.72
    \\
    RelationNet$^\dagger$~\cite{sung2018learning}&57.71$\pm$0.73
    \\
    CloserLook~\cite{chen2019closer}&65.57$\pm$0.70
    \\
    CloserLook++~\cite{chen2019closer}&62.04$\pm$0.76
    \\
    Centroid~\cite{afrasiyabi2020associative}&70.37$\pm$1.02
    \\
    Rethink-distill~\cite{tian2020rethinking}&68.57$\pm$0.39
    \\ \rowcolor{gray!20}
    \textbf{BML}&\textbf{72.42}$\pm$\textbf{0.54}
    \\ \thickhline
    \end{tabular}

\end{table}

\textbf{Cross domain evaluation} results is reported in Table~\ref{Tab:mini2CUB} (Results with $^\dagger$ are reported in \cite{afrasiyabi2020associative}), BML outperforms all compared methods with a large margin. Among them, the method in~\cite{tian2020rethinking} which also employs online soft label to improve the learning of inter-class relations, is 3.85\% lower than our binocular mutual learning mechanism. This shows that bidirectional cross-view interaction is better than unidirectional same-view interaction.

\input{Sec/Sec04/4_3_ablation_study}

%% file: Sec/Sec04/4_1_imagenet_serious.tex
\begin{table*}[t]
    \centering
    \caption{Comparison on \emph{mini}ImageNet, \emph{tiered}ImageNet and CIFAR-FS. Results with $^\dagger$ are reported in \cite{lee2019meta}. }
    \label{Tab:ImageNetserious}
    \vspace{2pt}
    \small
    \renewcommand\tabcolsep{7.5pt} 
    \begin{threeparttable}
    \begin{tabular}{lccccccc}
    \thickhline
    \multirow{2}{*}{\textbf{Method}}&\multirow{2}{*}{\textbf{Backbone}}&\multicolumn{2}{c}{\textbf{\emph{mini}ImageNet}}&\multicolumn{2}{c}{\textbf{\emph{tiered}ImageNet}}&\multicolumn{2}{c}{\textbf{CIFAR-FS}}
    \\ \cline{3-8}
    &&\emph{\textbf{1-shot}}&\emph{\textbf{5-shot}}&\emph{\textbf{1-shot}}&\emph{\textbf{5-shot}}&\emph{\textbf{1-shot}}&\emph{\textbf{5-shot}}
    \\ \thickhline
    \textbf{MAML}$^\dagger$~\cite{finn2017model}&\emph{ConvNet}&48.70$\pm$1.84&63.11$\pm$0.92&51.67$\pm$1.81&70.30$\pm$1.75&58.90$\pm$1.90&71.50$\pm$1.00
    \\ 
    \textbf{TAML}~\cite{jamal2019task}&\emph{ConvNet}&51.77$\pm$1.86&65.60$\pm$0.93&-&-&-&-
    \\ 
    \textbf{MetaOptNet}~\cite{lee2019meta}&\emph{ResNet12}&64.09$\pm$0.62&80.00$\pm$0.45&65.81$\pm$0.74&81.75$\pm$0.53&72.00$\pm$0.70&84.20$\pm$0.50
    \\ \thickhline
    \textbf{ProtoNet}$^\dagger$~\cite{snell2017prototypical}&\emph{ConvNet}&49.42$\pm$0.78&68.20$\pm$0.66&53.31$\pm$0.89&72.69$\pm$0.74&55.50$\pm$0.70&72.00$\pm$0.60
    \\ 
    \textbf{MatchingNet}$^\dagger$~\cite{vinyals2016matching}&\emph{ConvNet}&43.56$\pm$0.84&55.31$\pm$0.73&-&-&-&-
    \\ 
    \textbf{RelationNet}$^\dagger$~\cite{sung2018learning}&\emph{ConvNet}&50.44$\pm$0.82&65.32$\pm$0.70&54.48$\pm$0.93&71.32$\pm$0.78&55.00$\pm$1.00&69.30$\pm$0.80
    \\ 
    \textbf{DeepEMD}~\cite{zhang2020deepemd}&\emph{ResNet12}&65.91$\pm$0.82&82.41$\pm$0.56&71.16$\pm$0.80&86.03$\pm$0.58&46.47$\pm$0.70&63.22$\pm$0.71
    \\ 
    \textbf{FEAT}~\cite{ye2020few}&\emph{ResNet12}&66.78$\pm$0.20&82.05$\pm$0.14&70.80$\pm$0.23&84.79$\pm$0.16&-&-
    \\ 
    \textbf{TADAM}~\cite{oreshkin2018tadam}&\emph{ResNet12}&58.50$\pm$0.30&76.70$\pm$0.30&-&-&-&-
    \\ 
    \textbf{CTM}~\cite{li2019finding}&\emph{ResNet18}&64.12$\pm$0.82&80.51$\pm$0.13&68.41$\pm$0.39&84.28$\pm$1.73&-&-
    \\
    \textbf{LR+ICI}~\cite{wang2020instance}&\emph{ResNet12}&66.80$\pm$n/a&79.26$\pm$n/a&80.79$\pm$n/a&87.92$\pm$n/a&73.97$\pm$n/a&84.13$\pm$n/a
    \\ \thickhline
    \textbf{Rethink-Distill}~\cite{tian2020rethinking}&\emph{ResNet12}&64.82$\pm$0.60&82.14$\pm$0.43&71.52$\pm$0.69&86.03$\pm$0.49&73.90$\pm$0.80&86.90$\pm$0.50
    \\
    \textbf{DC}~\cite{lifchitz2019dense}&\emph{ResNet12}&61.26$\pm$0.20&79.01$\pm$0.13&-&-&-&-
    \\
    \textbf{MTL}~\cite{sun2019meta}&\emph{ResNet12}&61.20$\pm$1.80&75.50$\pm$0.80&-&-&-&-
    \\
    \textbf{CloserLook++}~\cite{chen2019closer}&\emph{ResNet18}&51.87$\pm$0.77&75.68$\pm$0.63&-&-&-&-
    \\
    \textbf{Meta-Baseline}~\cite{chen2020new}&\emph{ResNet12}&63.17$\pm$0.23&79.26$\pm$0.17&68.62$\pm$0.27&83.29$\pm$0.18&-&-
    \\
    \textbf{Neg-Cosine}~\cite{liu2020negative}&\emph{ResNet12}&63.85$\pm$0.81&81.57$\pm$0.56&-&-&-&-
    \\
    \textbf{AFHN}~\cite{li2020adversarial}&\emph{ResNet18}&62.38$\pm$0.72&78.16$\pm$0.56&-&-&68.32$\pm$0.93&81.45$\pm$0.87
    \\
    \textbf{Centroid}~\cite{afrasiyabi2020associative}&\emph{ResNet18}&59.88$\pm$0.67&80.35$\pm$0.73&69.29$\pm$0.56&85.97$\pm$0.49&-&-
    \\ \thickhline 
    \textbf{Baseline-\emph{local}}&\emph{ResNet12}&58.96$\pm$0.45&77.07$\pm$0.34&64.46$\pm$0.51&82.21$\pm$0.36&67.60$\pm$0.49&84.78$\pm$0.34
    \\
    \textbf{Baseline-\emph{global}}&\emph{ResNet12}&61.71$\pm$0.48&81.21$\pm$0.32&63.27$\pm$0.52&82.22$\pm$0.36&69.74$\pm$0.49&87.37$\pm$0.34
    \\ \rowcolor{gray!20}
    \textbf{BML}&\emph{ResNet12}&\textbf{67.04$\pm$0.63}&\textbf{83.63$\pm$0.29}&\textbf{68.99$\pm$0.50}&\textbf{85.49$\pm$0.34}&\textbf{73.45$\pm$0.47}&\textbf{88.04$\pm$0.33}
    \\ \thickhline
    \end{tabular}
    \end{threeparttable}
\end{table*}

%% file: Sec/Sec04/4_0_binocular_ablation.tex
\begin{figure*}[ht]
\vspace{-.5em}
    \centering
    \subfigure[1-shot]{
    \label{BML1}
    \includegraphics[width=0.23\linewidth]{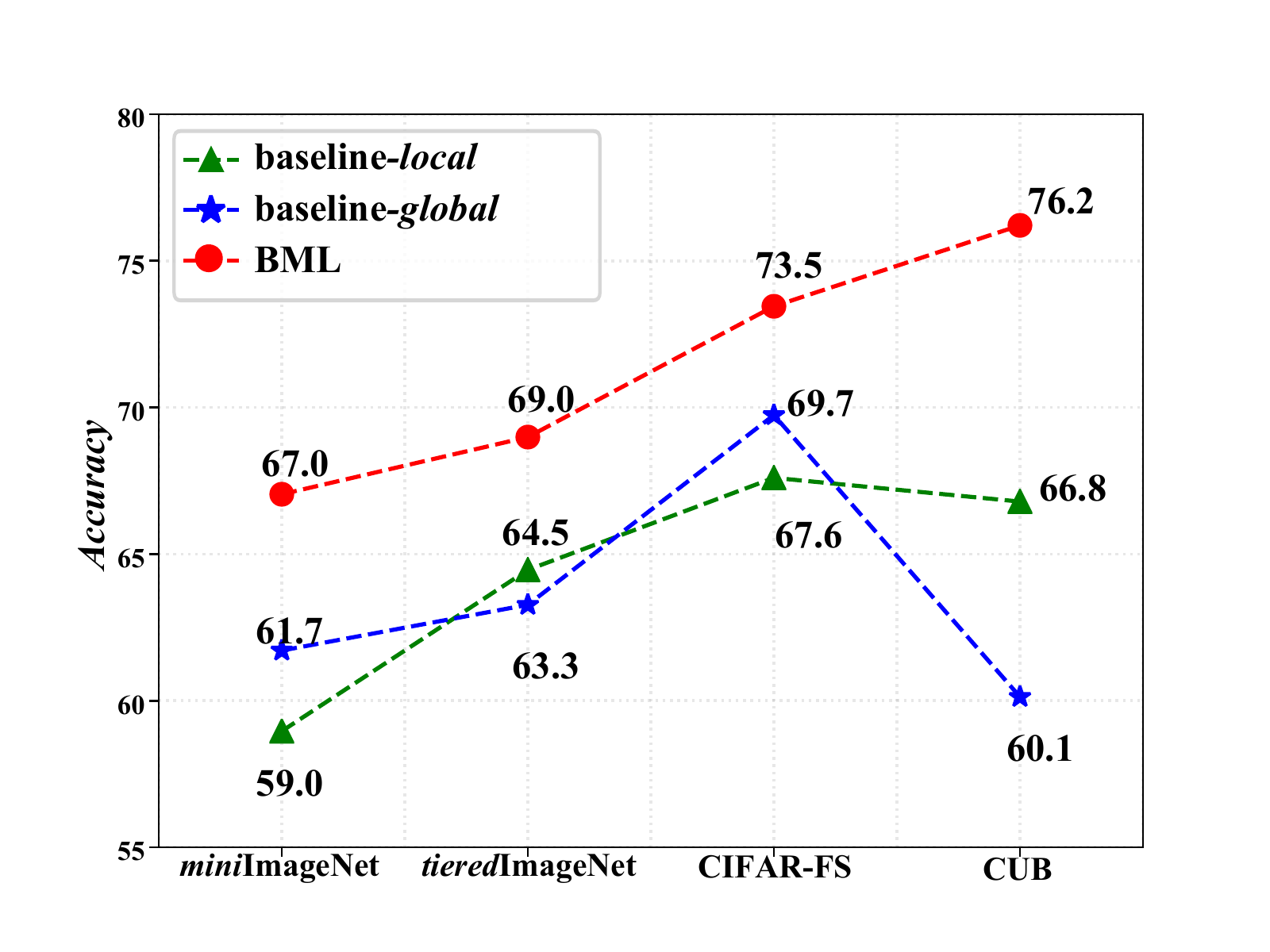}
    }
    \subfigure[5-shot]{
    \label{BML2}
    \includegraphics[width=0.23\linewidth]{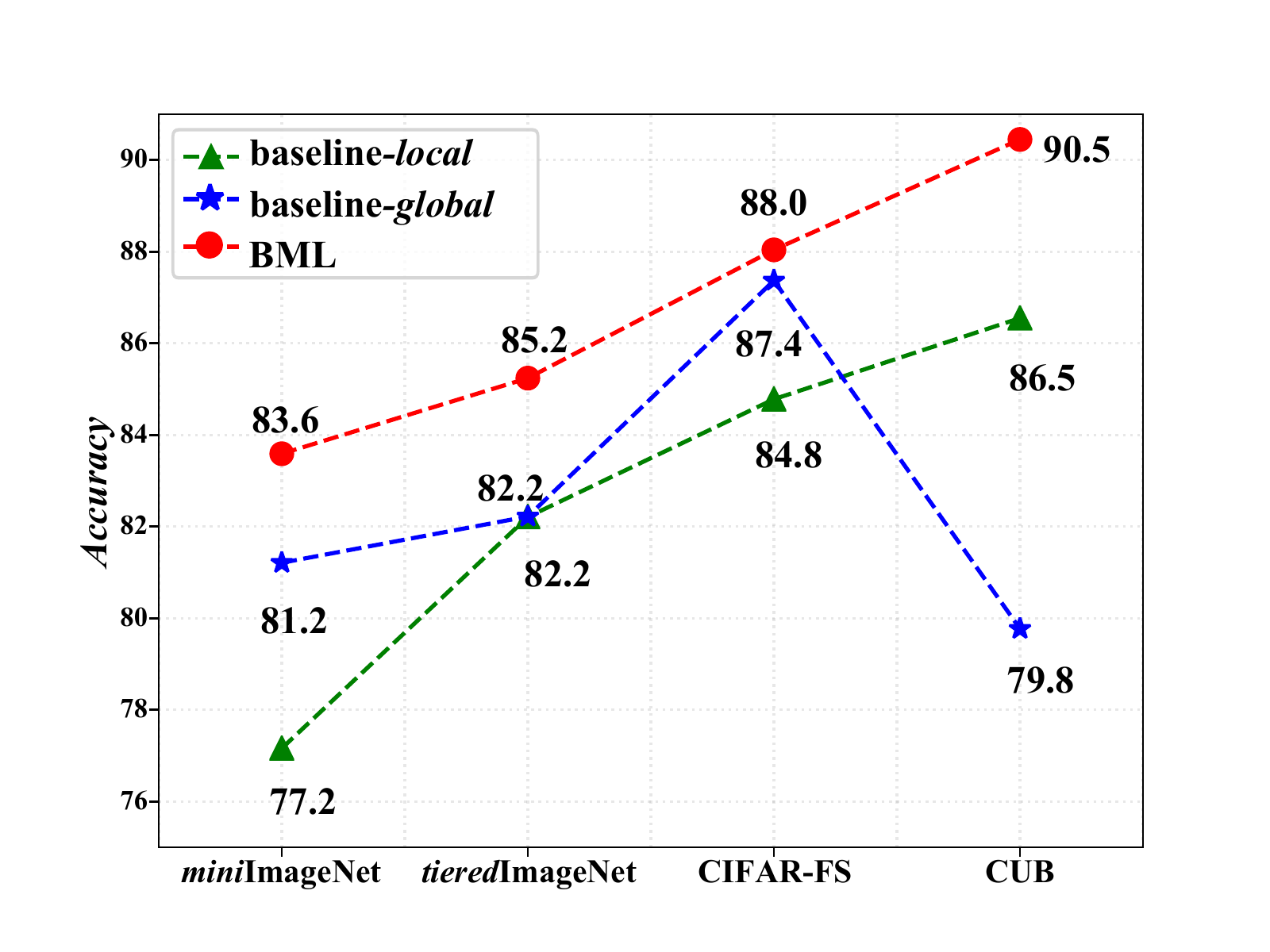}
    }
    \subfigure[1-shot]{
    \label{BML3}
    \includegraphics[width=0.23\linewidth]{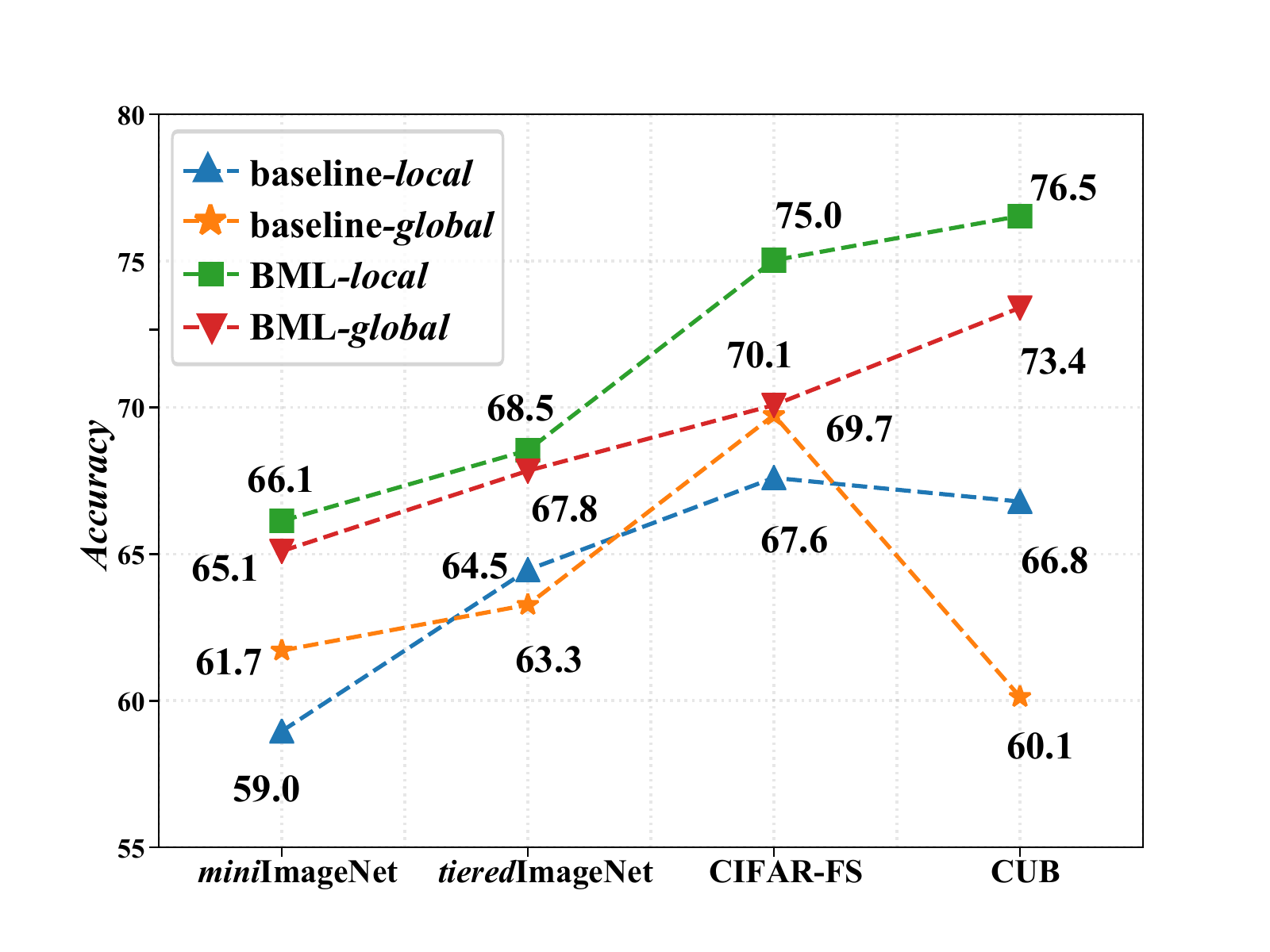}
    }
    \subfigure[5-shot]{
    \label{BML4}
    \includegraphics[width=0.23\linewidth]{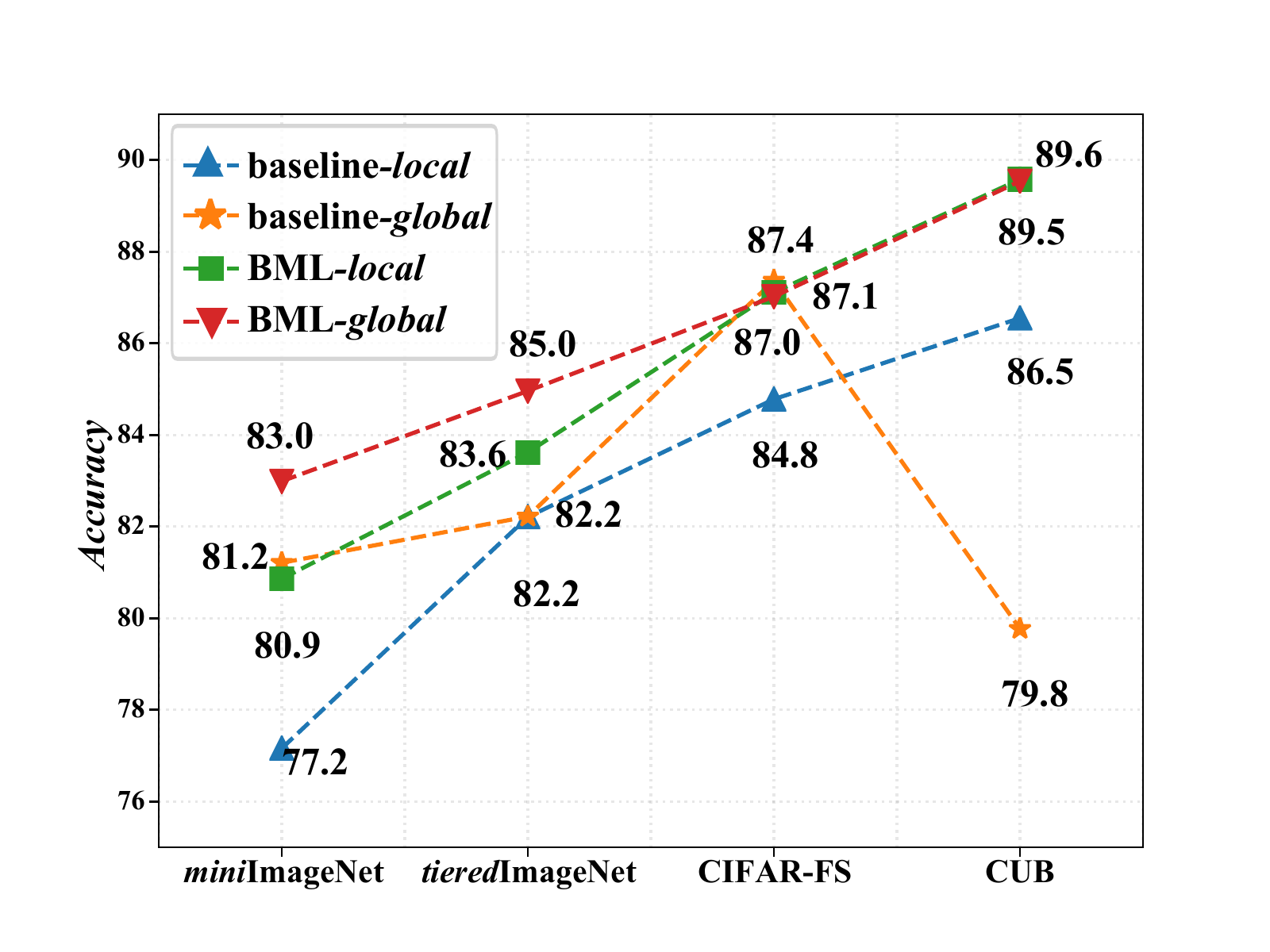}
    }
    \caption{Effect of binocular learning on multiple benchmarks. The comparison result proves the superiority of our BML.}
    \label{Fig:shots_ablation}
\end{figure*}

%% file: Sec/Sec04/4_3_ablation_study.tex
\subsubsection{Ablation Study}\label{sec4_2}
\noindent\textbf{Analyze of Binocular Learning:}
Specifically, our baselines are two typical single-view methods: baseline-\emph{global} and a stronger \emph{ProtoNet} baseline-\emph{local}. All the experimental configurations of the above two are the same as BML. According to the results shown in Figure~\ref{Fig:shots_ablation}, we highlight three observations:

1) Binocular mutual training can effectively integrate the advantages of the two views and obtain complementary integrated features. For instance, under 1-shot and 5-shot setting shown in subfigure.\ref{BML1} and \ref{BML2}, BML is significantly better than the two single-view training methods on both coarse-grained and fine-grained benchmarks.

2) Binocular mutual training makes the two views promote each other. As is shown in subfigure~\ref{BML3} and \ref{BML4}, BML-\emph{local} and BML-\emph{global} based on binocular training are better than baseline-\emph{local} and baseline-\emph{global} based on single-view mode, where BML-\emph{local} is 1.4\%-9.7\% higher than baseline-\emph{local}, and BML-\emph{global} is 1.8\%-13.3\% higher than baseline-\emph{global}. Since the only difference is whether the training is under binocular pattern, the results confirm the effectiveness of binocular training.

3) Binocular mutual training is less sensitive to shot count. By comparison, we find that BML-\emph{local} and BML-\emph{global} show different advantages under different settings. The \emph{global} branch performs better when shot count is greater than 1, indicating that \emph{global} view captures richer expressions; while under \emph{1-shot} setting, the \emph{local} branch exhibits a higher advantage, which shows that \emph{local} view is robust against sampling uncertainty, because local view minimizes the variation of features within a class according to the variation between classes~\cite{goldblum2020unraveling}. Binocular aggregation avoids the instability of single view and absorbs the advantages of two branches.

\noindent\textbf{Analyze of Mutual Interaction ($\mathcal{L}_{mutual}$):}
Applying mutual interaction loss $\mathcal{L}_{mutual}$ on the binocular framework, the performance is further improved by 0.79\% as follows:
{
\begin{center}
\vspace{-.2em}
\small
\begin{tabular}{p{55pt}|p{50pt}p{50pt}}
& w/o $\mathcal{L}_{mutual}$ & w/ $\mathcal{L}_{mutual}$ \\ \thickhline
BML & 82.84 & 83.63($\uparrow 0.79$)
\end{tabular}
\vspace{-.2em}
\end{center}
}

Besides, we separately analyze the impact of $\mathcal{L}_{mutual}$ on single view. Taking the local branch as example, the performance comparison with or without $\mathcal{L}_{mutual}$ is as follows:
{
\begin{center}
\vspace{-.2em}
\small
\begin{tabular}{p{55pt}|p{50pt}p{50pt}}
& w/o $\mathcal{L}_{mutual}$ & w/ $\mathcal{L}_{mutual}$ \\ \thickhline
BML-\emph{Local} & 79.29 & 80.95($\uparrow 1.66$)
\end{tabular}
\vspace{-.2em}
\end{center}
}
\begin{figure}[h]
    \centering
    \includegraphics[width=0.35\columnwidth]{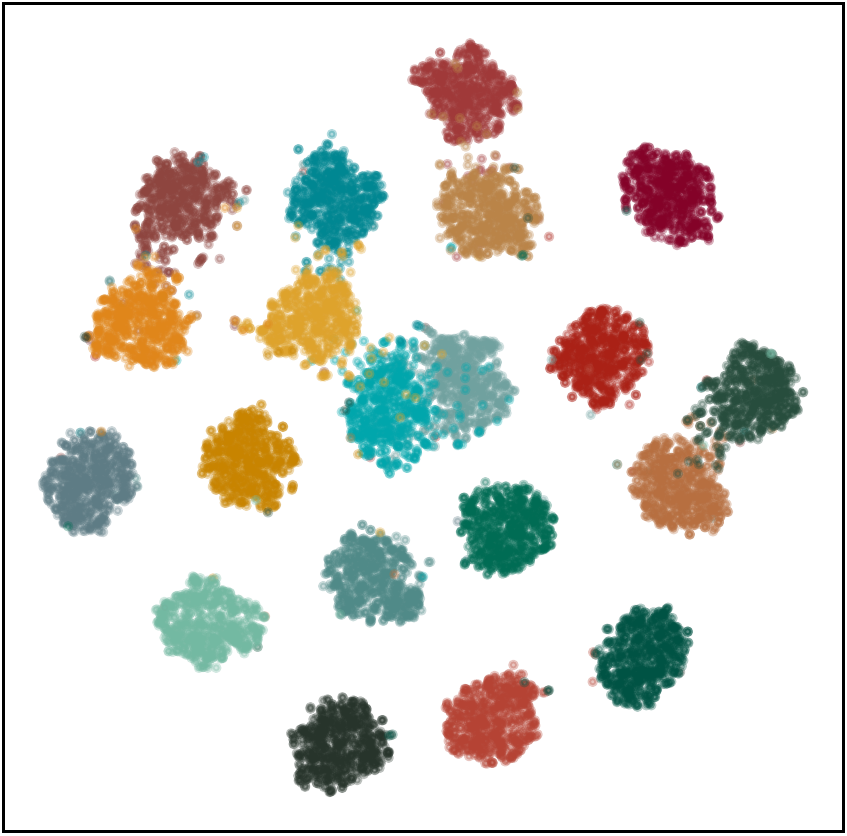}
    \includegraphics[width=0.35\columnwidth]{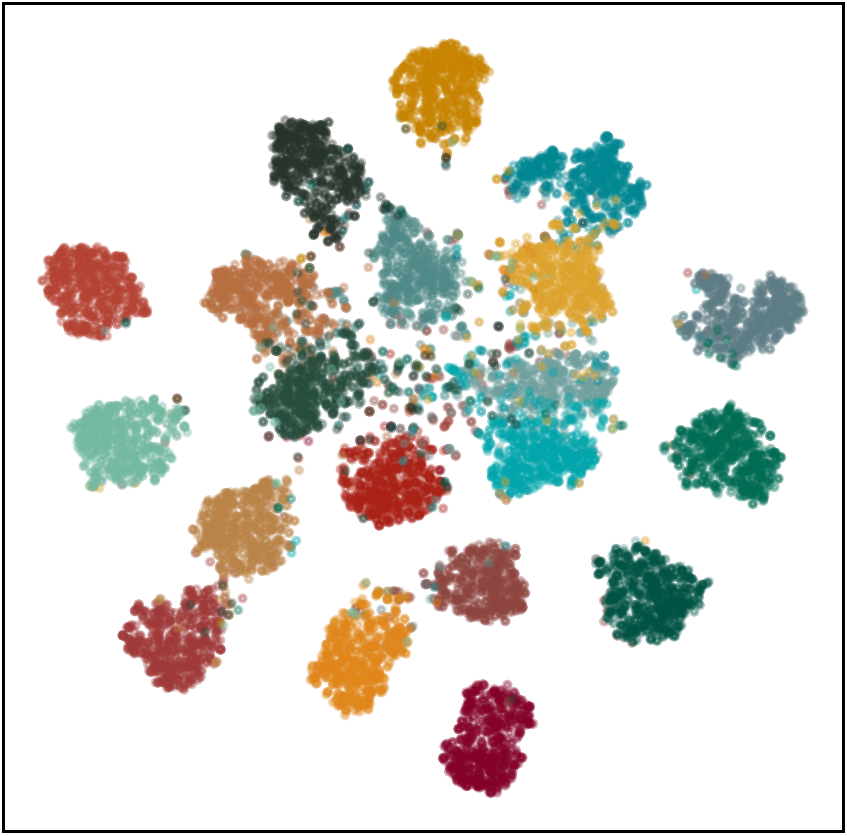}
    \caption{t-SNE results on \emph{mini}ImageNet base classes.}
    \label{Fig:MutualInt}
\end{figure}

Together with qualitative results in Figure~\ref{Fig:MutualInt}, we find that compared with the left one (w/o $\mathcal{L}_{mutual}$), after introducing $\mathcal{L}_{mutual}$ (right), the cluster structure on base classes is slightly broke but the performance on novel is significantly improved, which shows that in addition to the single hard label, minimizing the mimicry loss helps to alleviate the over-fitting problem while improving transferability.

\begin{figure}[h]
    \centering
    \includegraphics[width=0.8\columnwidth, height=0.45\columnwidth]{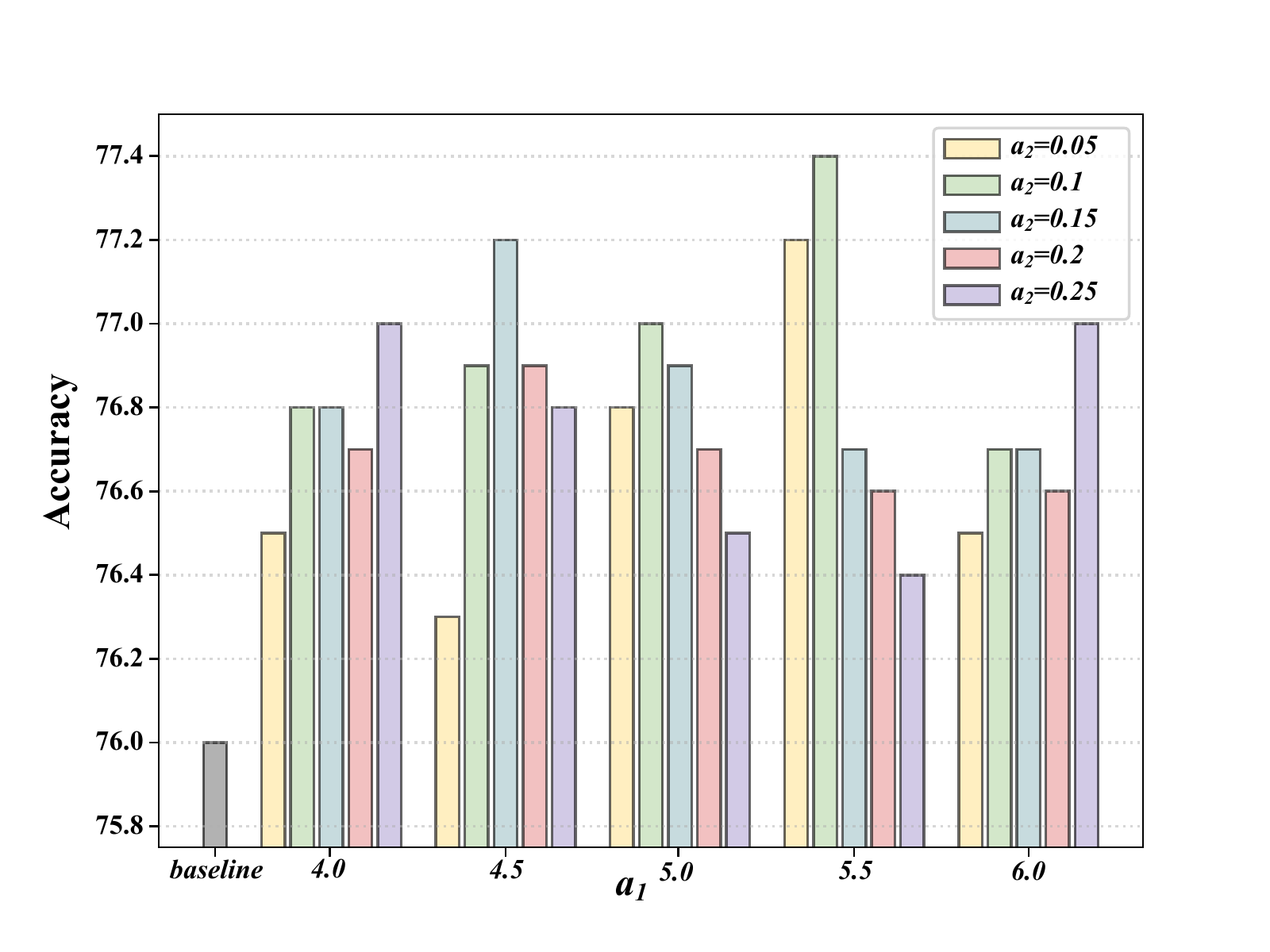}
    \caption{Effect of $\alpha_1$ and $\alpha_2$(train 100 epochs with $N$=5).}
    \label{Fig:ELL_ablation_a1_a2}
\end{figure}

\noindent\textbf{Analyze of Elastic Loss (ELloss):} We simply update the basic euclidean distance with the elastic loss, and keep the other settings unchanged. Here is the comparison result (all the experiments are conducted on \emph{mini}ImageNet):
{
\begin{center}
\vspace{-.2em}
\small
\begin{tabular}{p{55pt}|p{50pt}p{50pt}}
& w/o ELloss & w/ ELloss \\ \thickhline
baseline-\emph{Local} & 77.07 & 77.77($\uparrow 0.7$)
\end{tabular}
\vspace{-.2em}
\end{center}
}
Obviously, the introduction of ELloss has brought a great performance improvement. Next, we made a further detailed analysis:

\textbf{Effect of $\alpha_1$ and $\alpha_2$:} $\alpha_1$ and $\alpha_2$ in Algorithm.\ref{alg02} are two factors controlling the push degree. Among them, $\alpha_1\in$[4, 6], this range is derived from the observation of normal training: $\mathcal{M}(C_p, x_q)$ stably falls in [6, 12] after convergence. Therefore, we enlarge $\mathcal{M}(C_p, x_q)$ about 50\%. The range of $\alpha_2$ is [0.05, 0.25], which controls push degree cross tasks. As shown in Figure~\ref{Fig:ELL_ablation_a1_a2}, simply employing ELloss on baseline-\emph{local} can obtain a significant improvement (best configuration $\alpha_1=5.5$, $\alpha_2=0.1$), which shows that, compare to equally treats all the tasks, redefining the difficulty is beneficial to dig out more transferable knowledge.

\begin{figure}[t]
    \centering
    \includegraphics[width=0.48\linewidth]{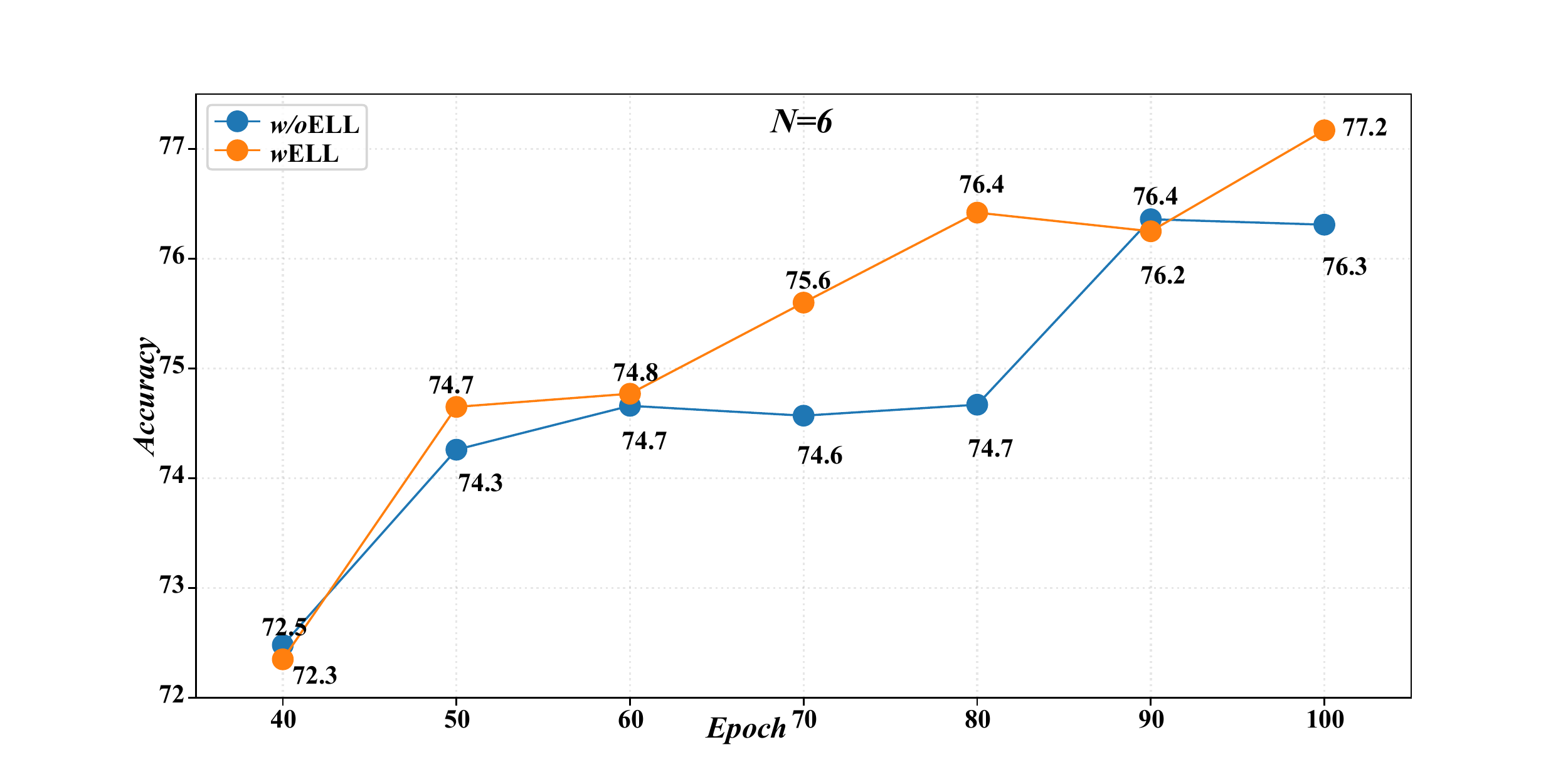}
    \includegraphics[width=0.48\linewidth]{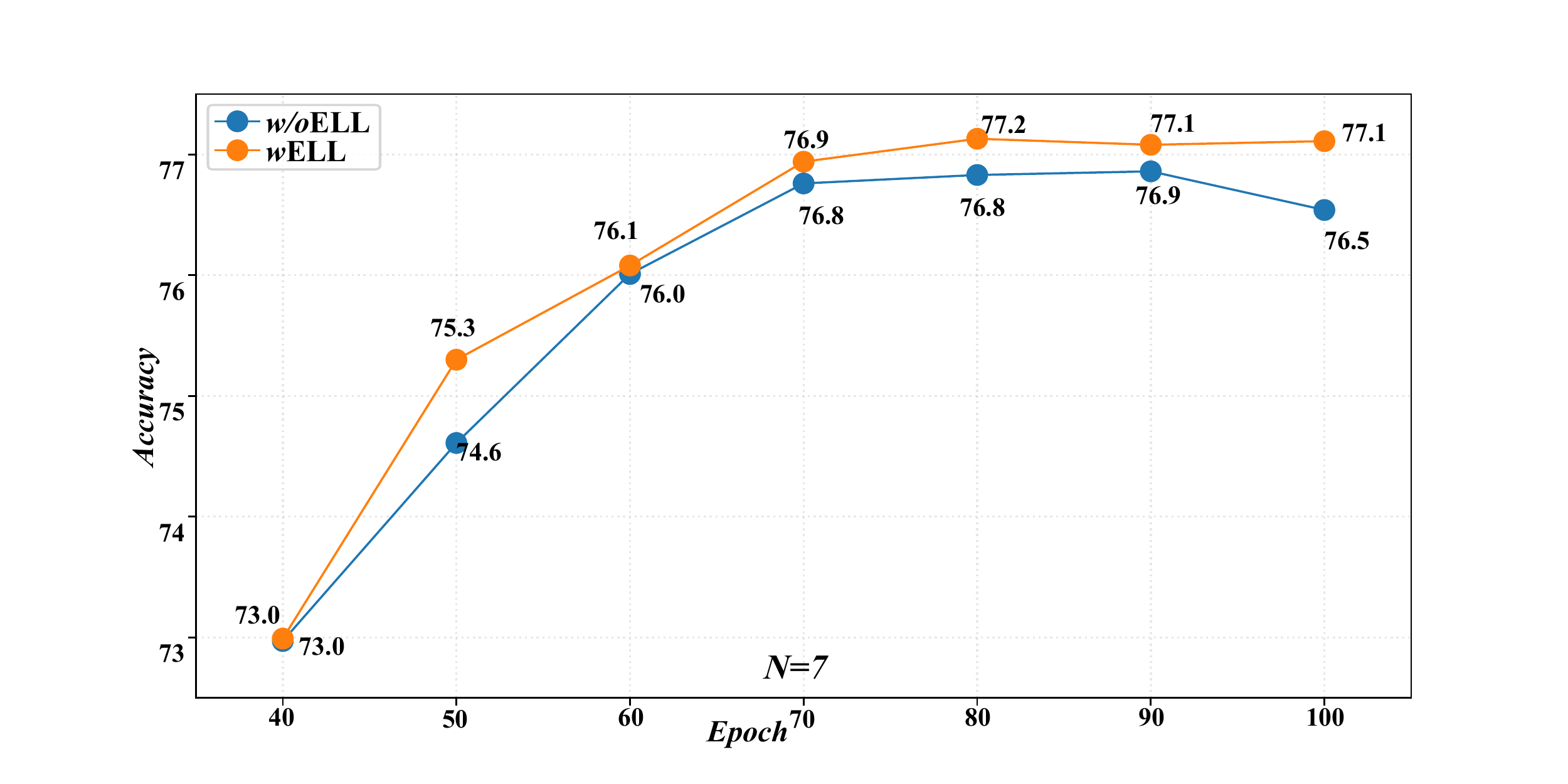}

    \includegraphics[width=0.48\linewidth]{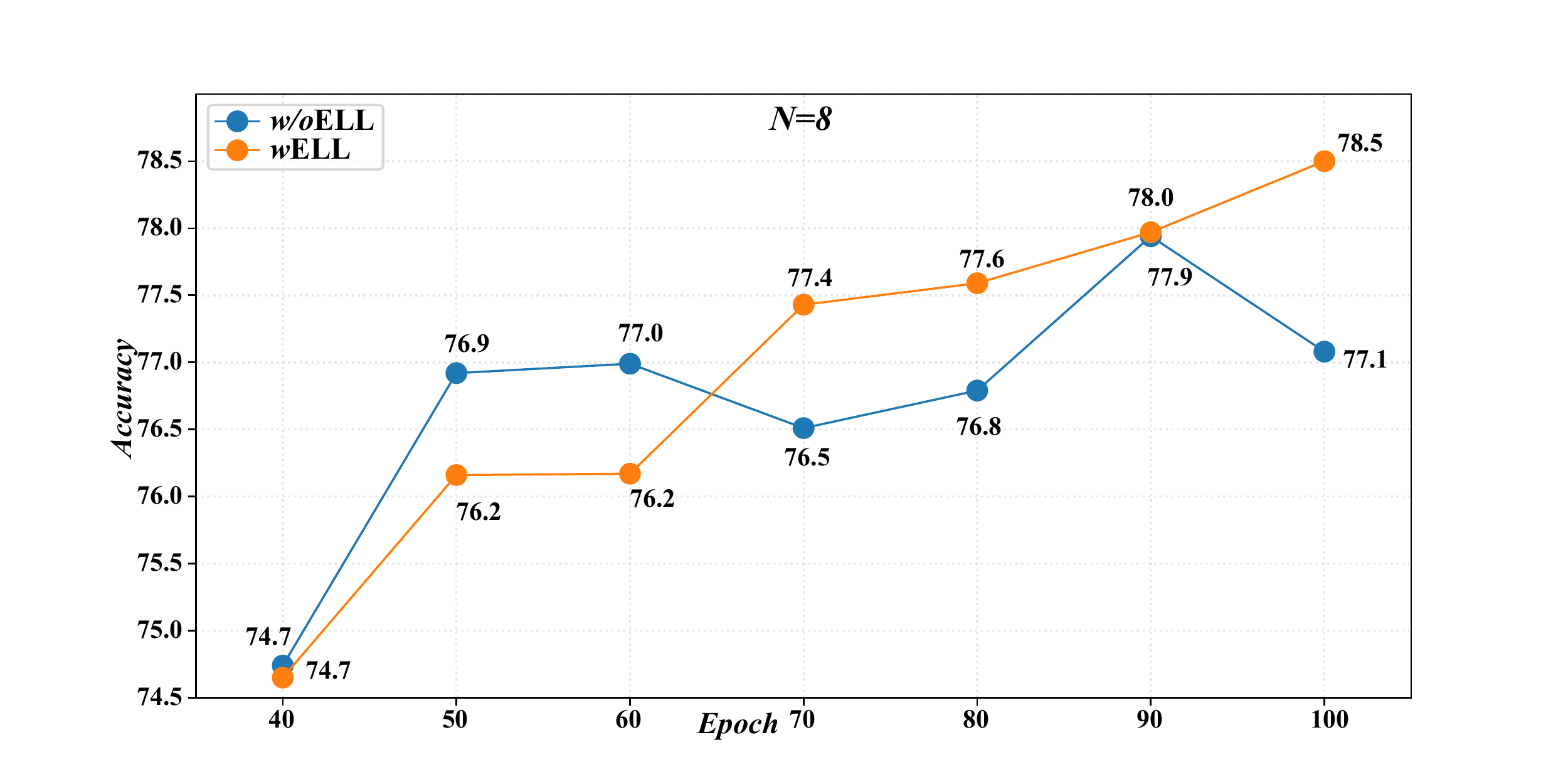}
    \includegraphics[width=0.48\linewidth]{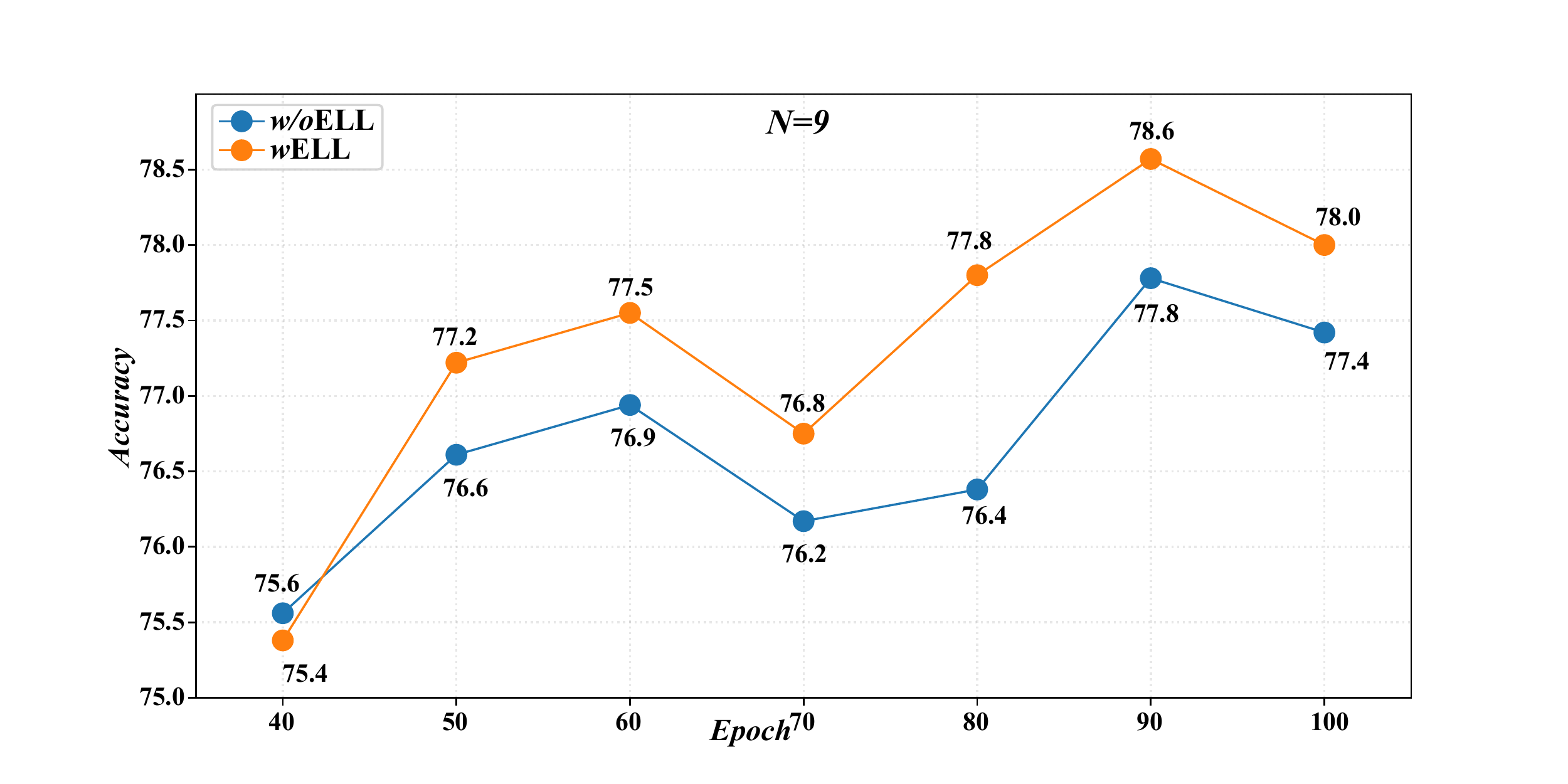}
    \caption{Performance trend of Elastic Loss with different $N$.}
    \label{Fig:ELL_ablation_N}
\end{figure}

\textbf{Effect of Elastic Loss under different settings:} We further analyze ELloss in Figure~\ref{Fig:ELL_ablation_N}, and the result shows a fact that local mode is indeed a relatively simple problem since the matching process only occurs within current episode, which naturally, makes the network too lazy to give satisfied solutions. The introduction of ELloss delays the appearance of performance saturation points, thereby alleviating the problem of sub-optimal solutions to some extent.

\noindent\textbf{Analyze of Stability:}
A good model should perform well in any situation, especially when attacked by degraded input. The following stability test in Table~\ref{Tab:stablity} is carried out in several groups of degraded cases: size change; blur attack; noise attack and brightness attack. Results show that our BML performs well in any situation, which shows the stability of binocular learning strategy.
\begin{table}[ht]
    \centering
    \caption{Stability evaluation on \emph{mini}ImageNet.}
    \label{Tab:stablity}
    \small
    \begin{tabular}{l|c}
    \thickhline
    &$84\times84 \to 224\times224$
    \\ \cline{2-2}
    DeepEMD~\cite{zhang2020deepemd}&82.41$\to$78.12
    \\ \rowcolor{gray!20}
    \textbf{BML}&\color{red}{\textbf{83.59$\to$81.57}}
    \\ \thickhline
    &$+$ GaussianBlur ($\sigma\in[0.1, 2]$)
    \\ \cline{2-2}
    Rethink-Distill~\cite{tian2020rethinking}&82.14$\to 49.30$
    \\ \rowcolor{gray!20}
    \textbf{BML}&\color{red}{\textbf{83.59$\to$61.96}}
    \\ \thickhline
    &$+$pepperNoise ($r=0.01$)
    \\ \cline{2-2}
    Rethink-Distill~\cite{tian2020rethinking}&82.14$\to$63.97
    \\ \rowcolor{gray!20}
    \textbf{BML}&\color{red}{\textbf{83.59$\to$71.02}}
    \\ \thickhline
    &$+$ ColorJitter ($B=0.8$)
    \\
    Rethink-Distill~\cite{tian2020rethinking}&82.14$\to$81.05
    \\ \rowcolor{gray!20}
    \textbf{BML}&\color{red}{\textbf{83.59$\to$82.24}}
    \\ \thickhline
    \end{tabular}
\end{table}

\begin{figure}[t]
   \centering
   \includegraphics[width=0.8\linewidth]{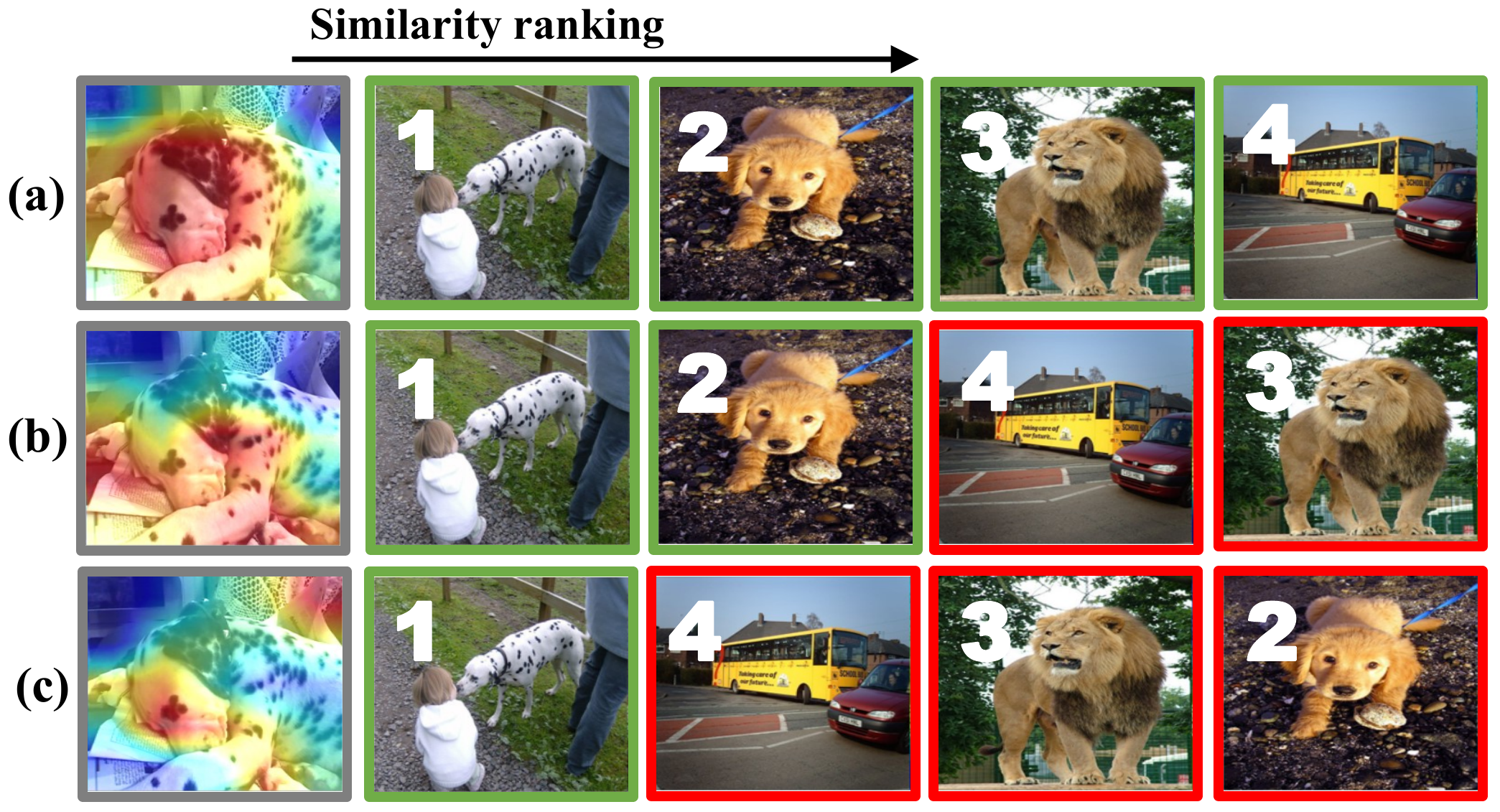}
   \caption{Similarity Ranking on a 4-way 1-shot task (\emph{golden retriever}, \emph{dalmatian}, \emph{lion} and \emph{bus}). (a) Our \textbf{BML}, (b)single \emph{global} view, (c)single \emph{local} view. BML gives accurate ranking results while the left two failed to get inter-class relationship.}
\label{fig05}
\end{figure}

\noindent\textbf{Is Similarity Ranking important for a good model?}
We randomly visualize a task (Figure~\ref{fig05}) and find an interesting sidelight: while improving discriminability, BML also gets more accurate inter-class relationship and more accurate heatmap. This inspired us to think: for the current closed-set setup of FSC, should we pay attention to the similarity ranking besides the top-1 accuracy? Since semantical effective ranking is more practical and can further distinguish the advantages of existing methods.

%% file: Sec/Supp/supp.tex
%%%%%%%%% BODY TEXT
\section{More Experimental Results}
\subsection{More Shots}
\begin{figure}[h]
    \centering
    \subfigure[10-shot]{
    \includegraphics[width=0.45\columnwidth]{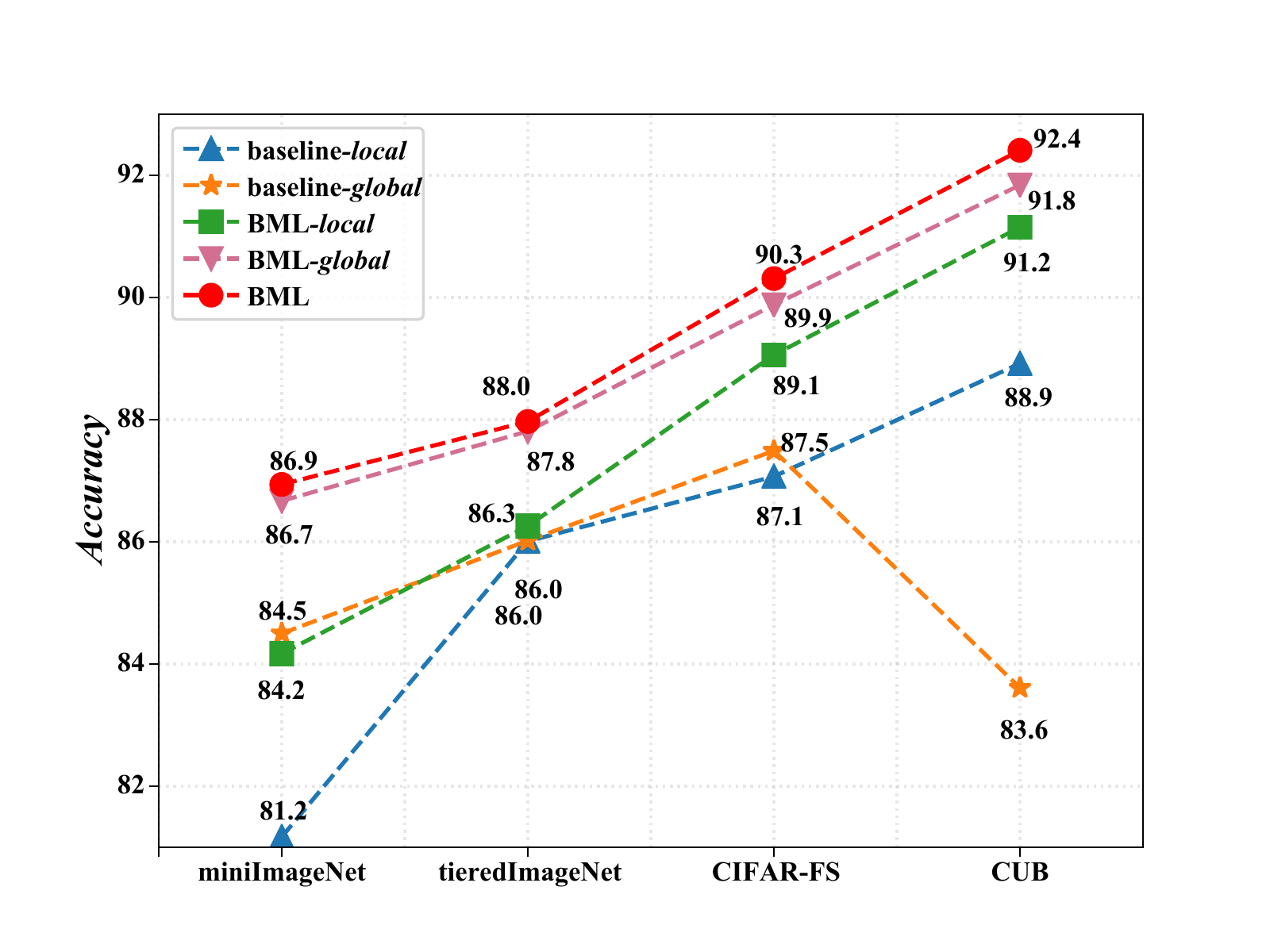}
    }
    \subfigure[20-shot]{
    \includegraphics[width=0.45\columnwidth]{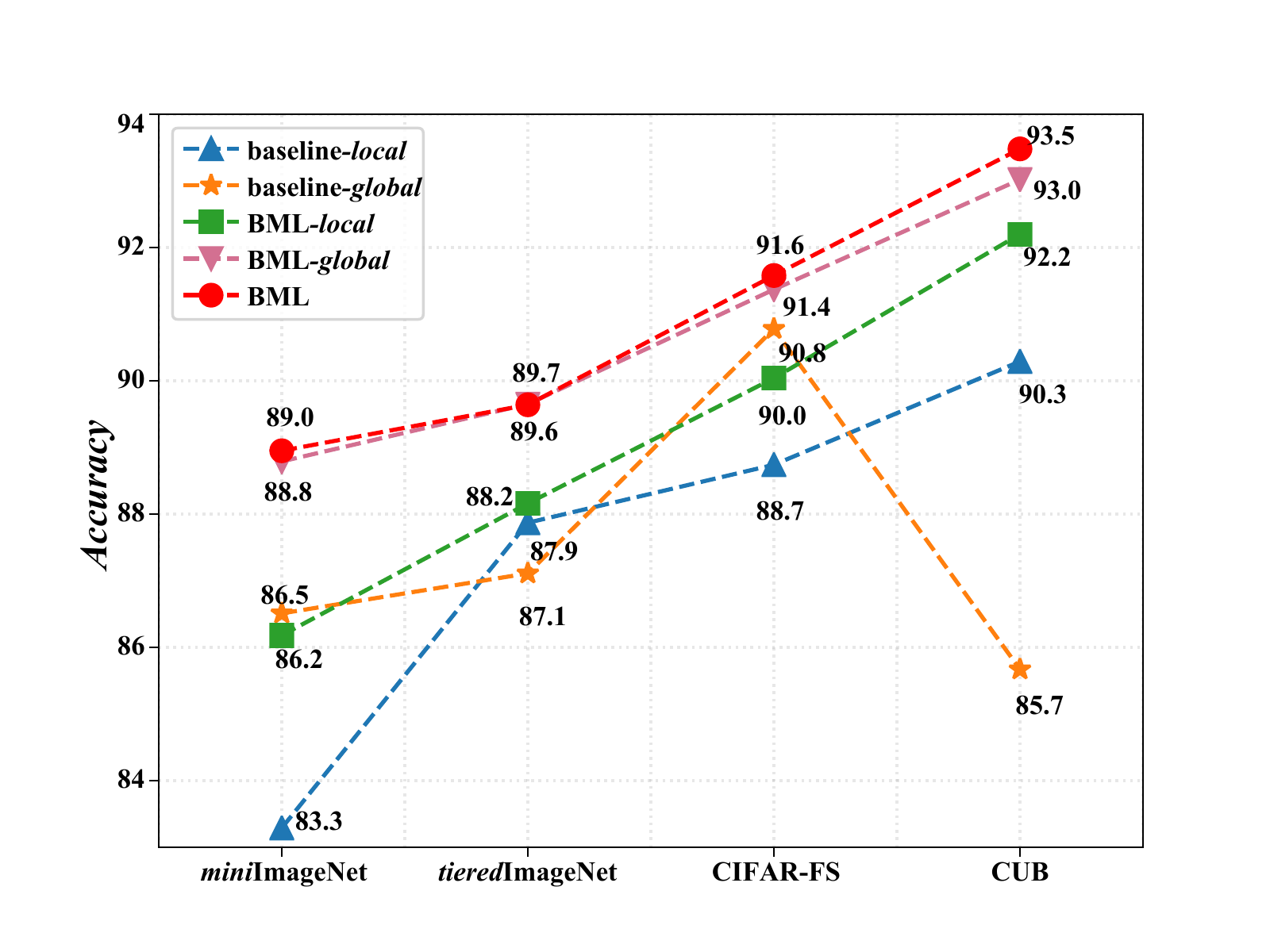}
    }
    \caption{Comparison between BML and the two baselines with more shots.}
    \label{Fig_1}
\end{figure}

As the number of visible samples (support shots) increases (Figure ~\ref{Fig_1}), performance gradually improves, and BML is steadily higher than the two single-view baselines. Besides, the performance of BML\emph{-global} and BML\emph{-local} under the binocular mode is superior to baseline\emph{-global} and baseline\emph{-local} under single view mode.
\subsection{More Benchmarks}
To further verify the performance of BML, we do experiments on another public few-shot classification benchmark: FC100. FC100 is derived from CIFAR-100, which has a total of 100 classes. Among them, 60 classes are used for training, 20 are used for verification, and the remaining 20 are used for testing. Since the division is carried out at the superclass level, the information overlap between splits is minimized, thus more challenging.

As is shown in Table~\ref{Tab:FC100}, on FC100, BML is still superior to the two single-view baselines, and stays ahead of the other six competitors. Specifically, three key points are conveyed which have been emphasized in Section 4:
\begin{itemize}[itemsep=-5pt,topsep=5pt]
    \item Binocular learning is better than single-view mode. BML is 2\% higher than the single global view and 5\%-9\% higher than the single local view.
    \item On coarse-grained dataset, global view performs better than local view.
    \item The two complementary views can promote each other (\emph{i.e.}, BML\emph{-global} vs. baseline\emph{-global}, BML\emph{-local} vs. baseline\emph{-local}), and the global impact on the local view is more obvious.
\end{itemize}
\begin{table}[h]
    \centering
    \caption{Comparison on FC100.}
    \label{Tab:FC100}
    \small
    \renewcommand\tabcolsep{15.0pt}
    \begin{threeparttable}
    \begin{tabular}{lcc}
    \thickhline
    \multirow{2}{*}{\textbf{Method}}&\multicolumn{2}{c}{\textbf{FC100}}
    \\ \cline{2-3}
    &\textbf{\emph{5-way 1-shot}}&\textbf{\emph{5-way 5-shot}}
    \\ \thickhline
    \textbf{MAML}&38.10$\pm$1.70&50.40$\pm$1.00
    \\
    \textbf{MetaOptNet}&41.10$\pm$0.60&55.50$\pm$0.60
    \\
    \textbf{ProtoNet}&35.30$\pm$0.60&48.60$\pm$0.60
    \\
    \textbf{TADAM}&40.10$\pm$0.40&56.10$\pm$0.40
    \\
    \textbf{Rethink}&42.60$\pm$0.70&59.10$\pm$0.60
    \\
    \textbf{DC}&42.04$\pm$0.17&57.05$\pm$0.16
    \\ \thickhline
    \textbf{Baseline-\emph{local}}&38.88$\pm$0.38&54.25$\pm$0.40
    \\
    \textbf{Baseline-\emph{global}}&42.61$\pm$0.39&61.03$\pm$0.40
    \\
    \textbf{BML-\emph{local}}&43.25$\pm$0.41&58.70$\pm$0.39
    \\
    \textbf{BML-\emph{global}}&43.88$\pm$0.40&62.06$\pm$0.39
    \\ \rowcolor{gray!20}
    \textbf{BML}&45.00$\pm$0.41&63.03$\pm$0.41
    \\ \thickhline
    \end{tabular}
    \end{threeparttable}
\end{table}
\subsection{More Analyze of elastic loss}
\begin{figure}[h]
    \centering
     \subfigure[\emph{w/o} elastic loss]{
       \label{noEL}
       \includegraphics[width=0.49\columnwidth]{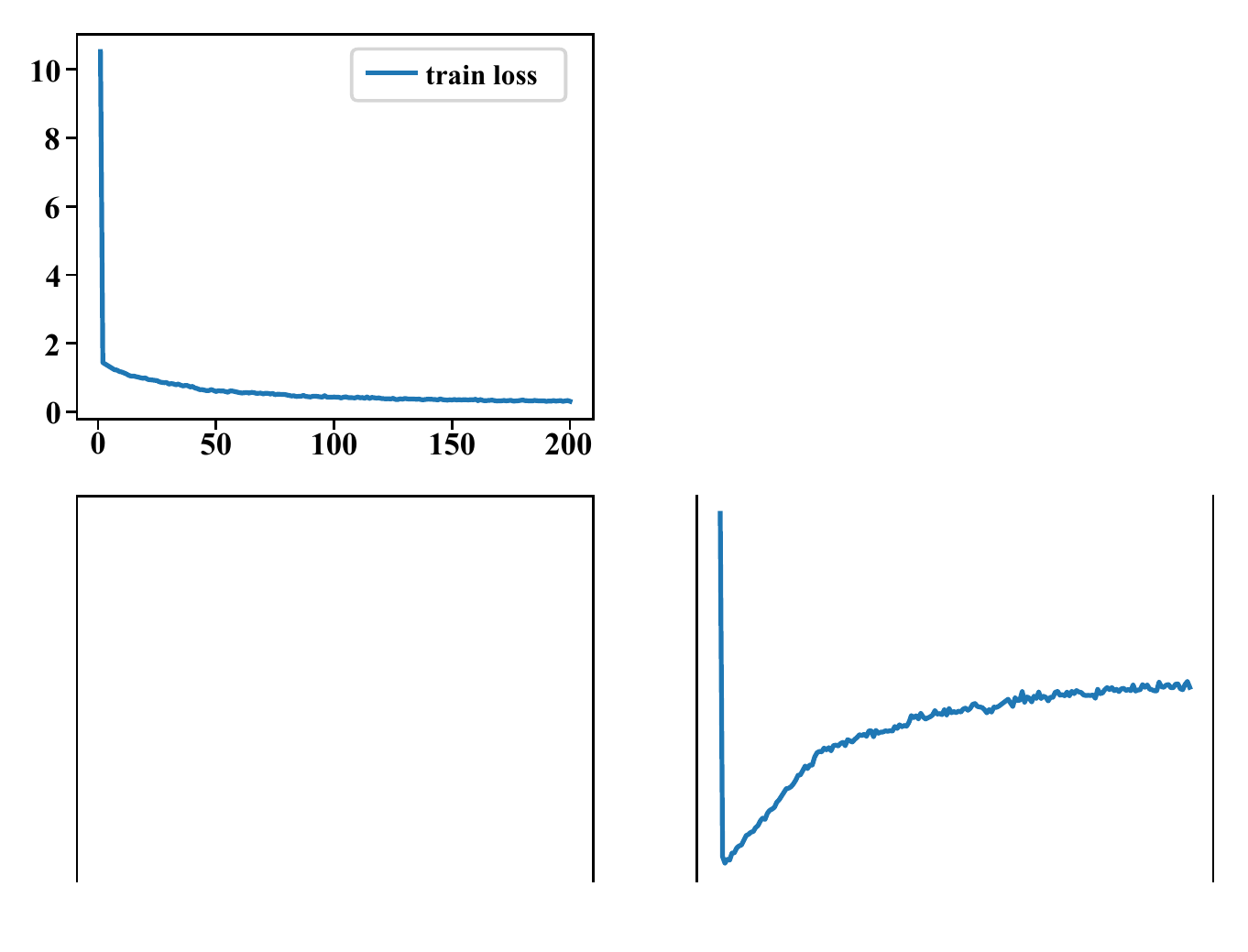}
       \includegraphics[width=0.475\columnwidth]{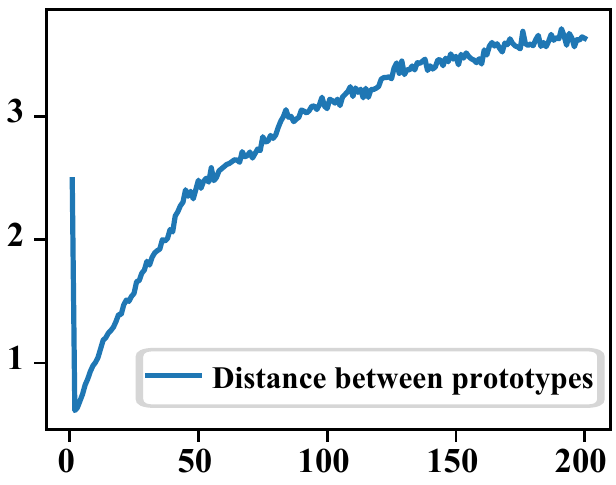}
    }
    \subfigure[\emph{w/} elastic loss]{
    \label{EL}
     \includegraphics[width=0.5\columnwidth]{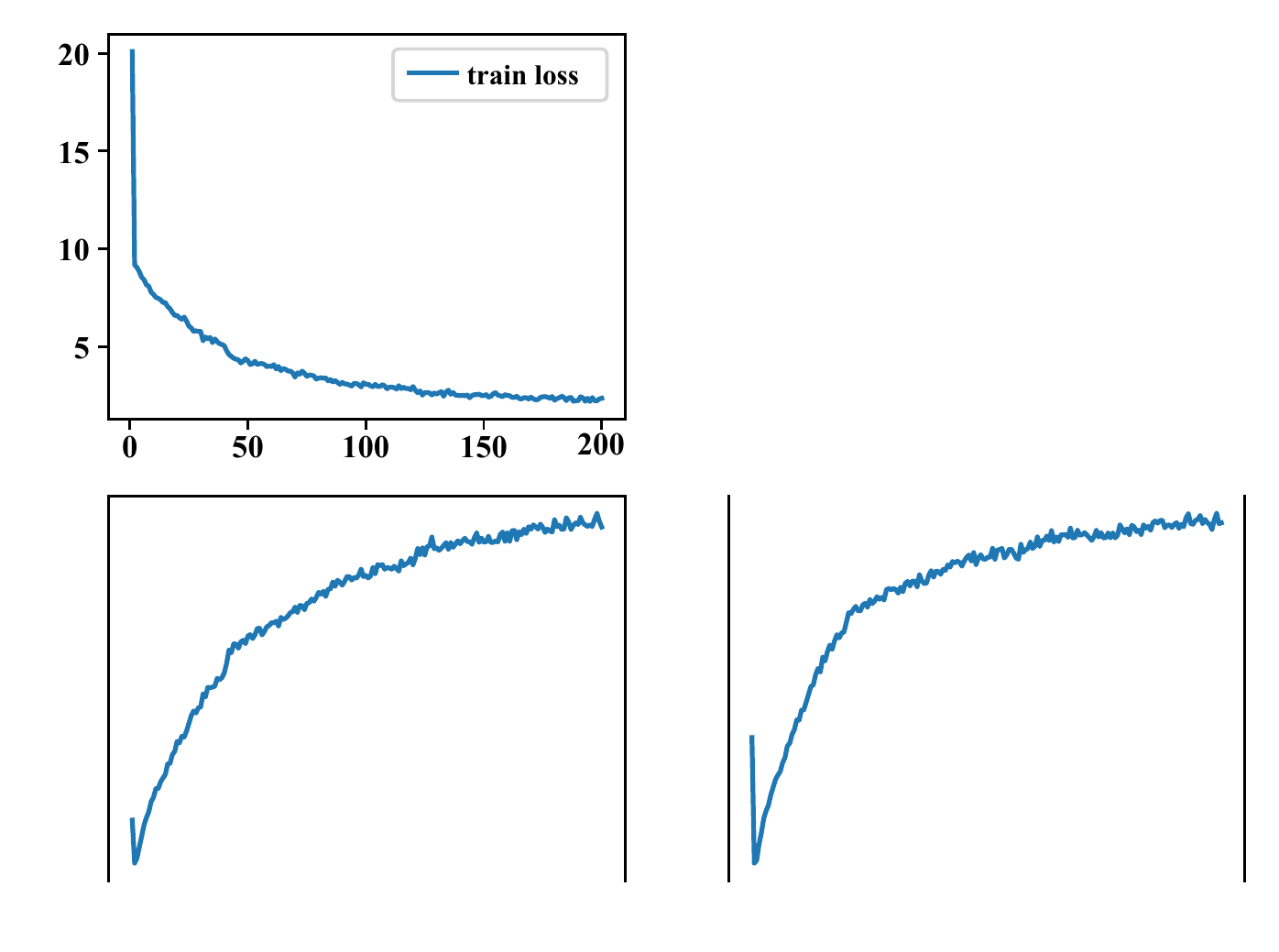}
     \includegraphics[width=0.49\columnwidth]{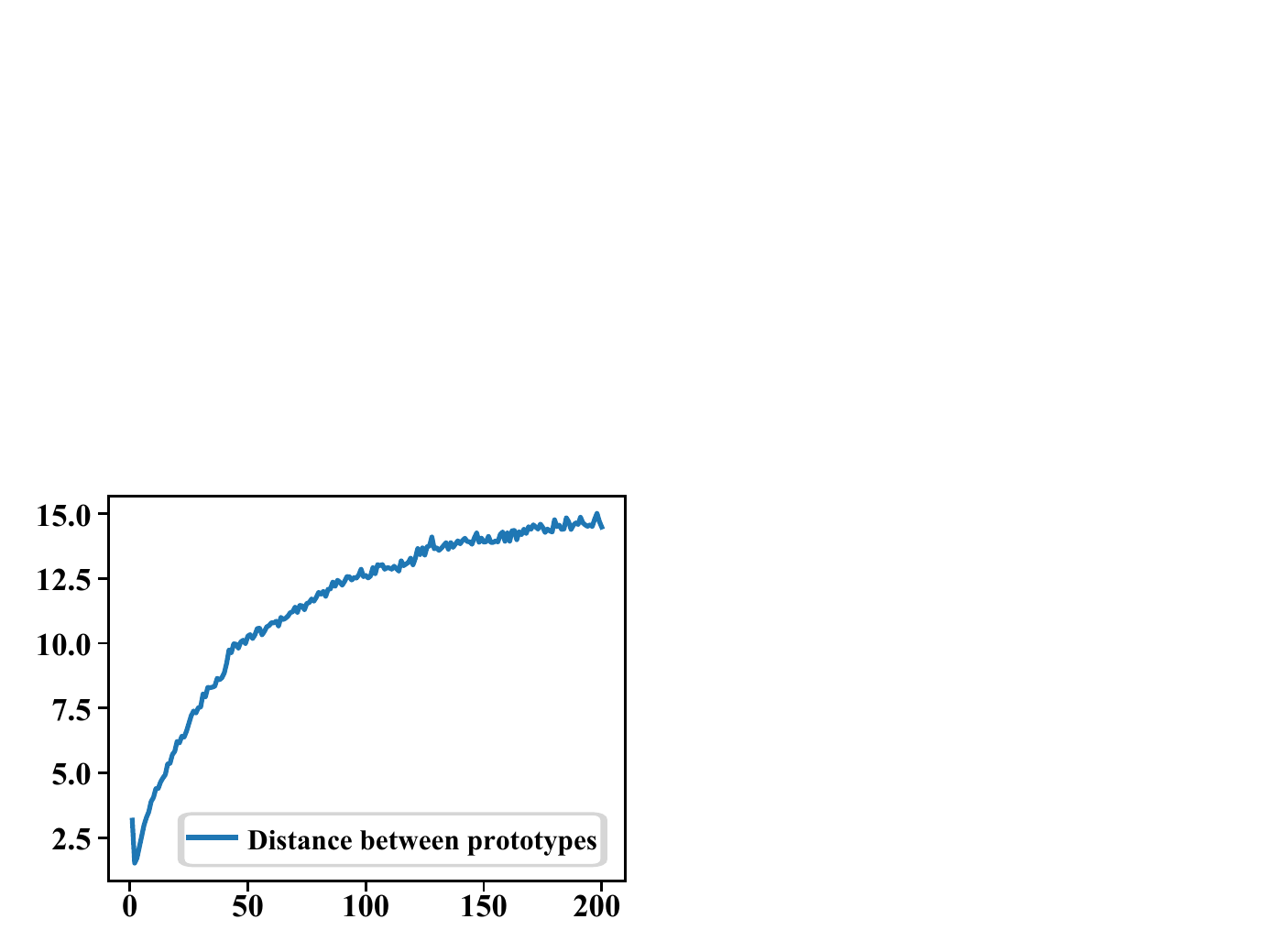}
    }
    \caption{Loss value and mean distance between prototypes on the base set.}
    \label{Fig_2}
\end{figure}

We carefully monitor the elastic loss and further explore its mechanism. Figure~\ref{Fig_2} shows the trend of \emph{training loss} and \emph{distance between prototypes} with or without elastic loss (On \emph{mini}ImageNet). Obviously, comparing the left subfigure of Figure~\ref{noEL} with the one of Figure~\ref{EL}, we can find that when no elastic loss is applied, the loss value quickly drops to a low point, and the subsequent decline has been very slow. On the contrary, after applying the elastic loss, the initial loss value is increased significantly, and the downward trend is more obvious. This shows that elastic loss does increase the difficulty of optimization. Furthermore, as shown in right subfigures of Figure~\ref{noEL} and Figure~\ref{EL}, the distance between $N$ prototypes (first-order moment) shows a similar change. With the help of elastic loss, the distance between prototypes is gradually expanded, and the features are more dispersed in the embedding space. This shows that the network is learning to amplify the difference between prototypes to improve matching accuracy.

\subsection{More Visualization of tasks}
\begin{figure}[ht]
    \centering
    \includegraphics[width=0.32\columnwidth]{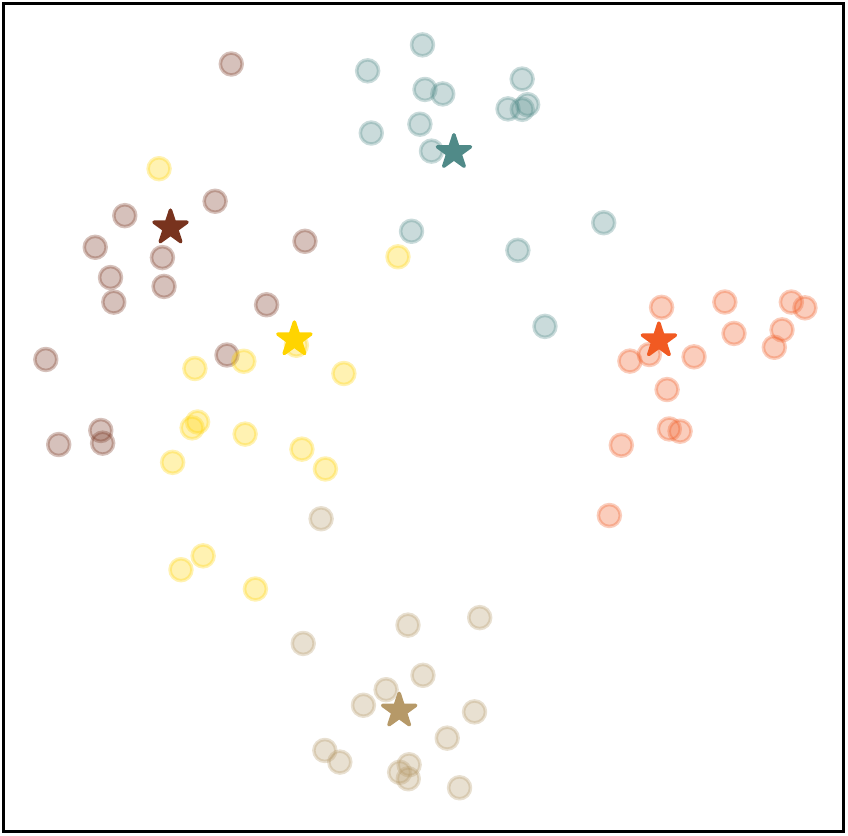}
    \includegraphics[width=0.32\columnwidth]{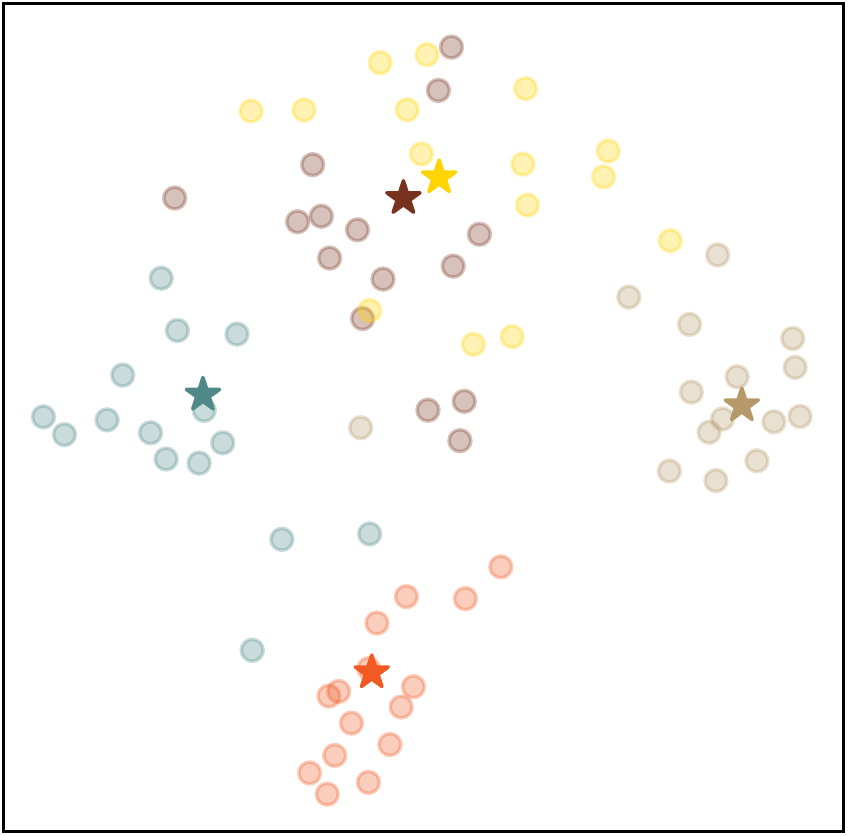}
    \includegraphics[width=0.32\columnwidth]{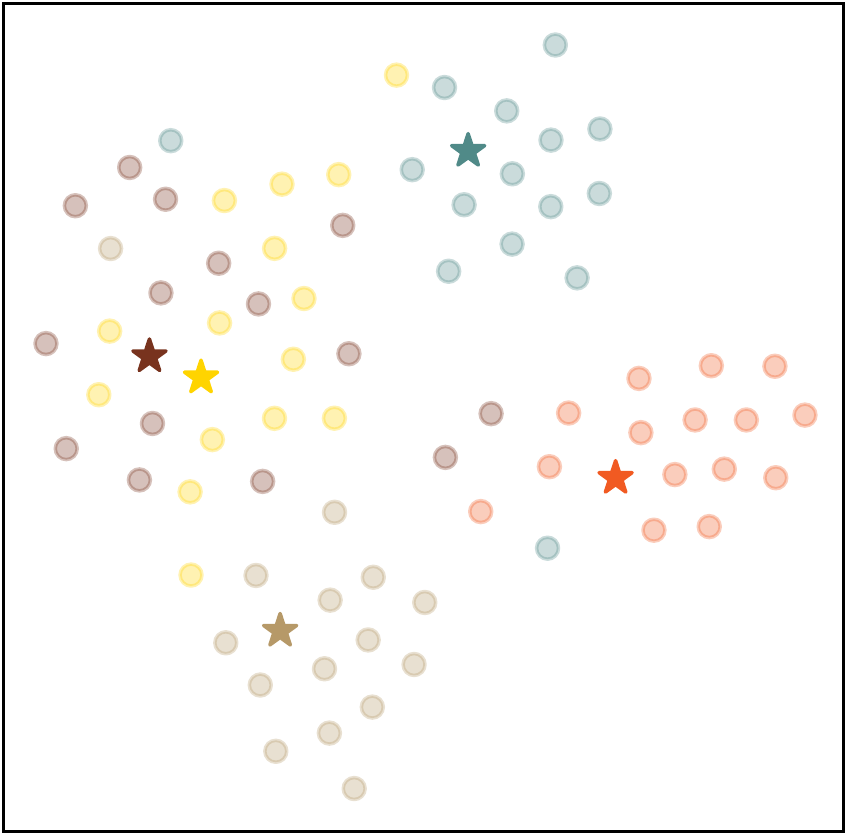}

    \includegraphics[width=0.32\columnwidth]{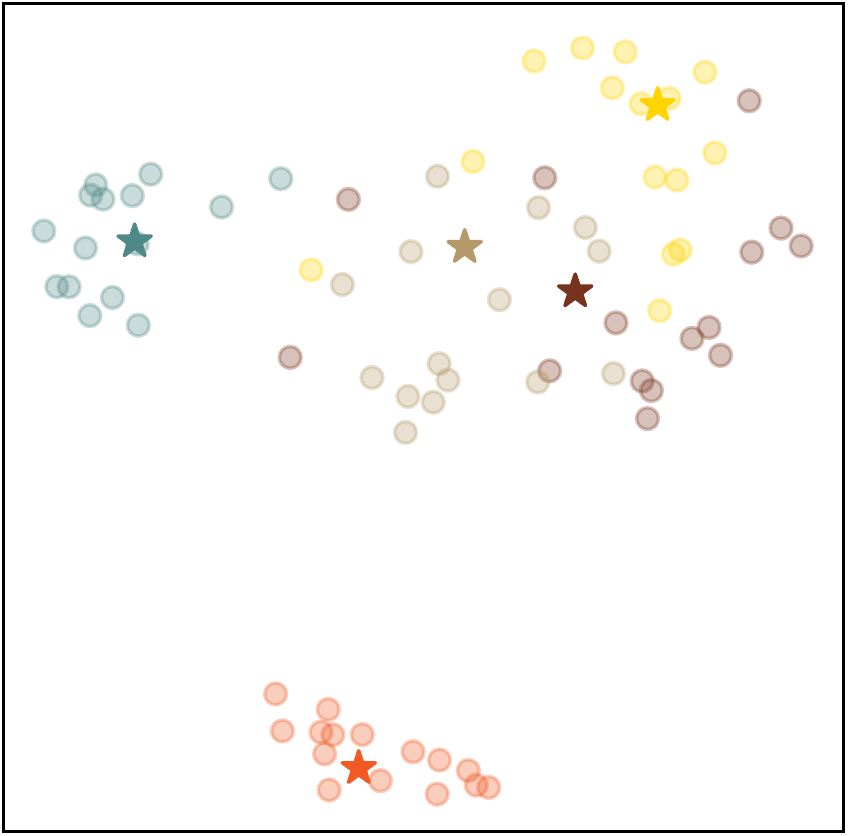}
    \includegraphics[width=0.32\columnwidth]{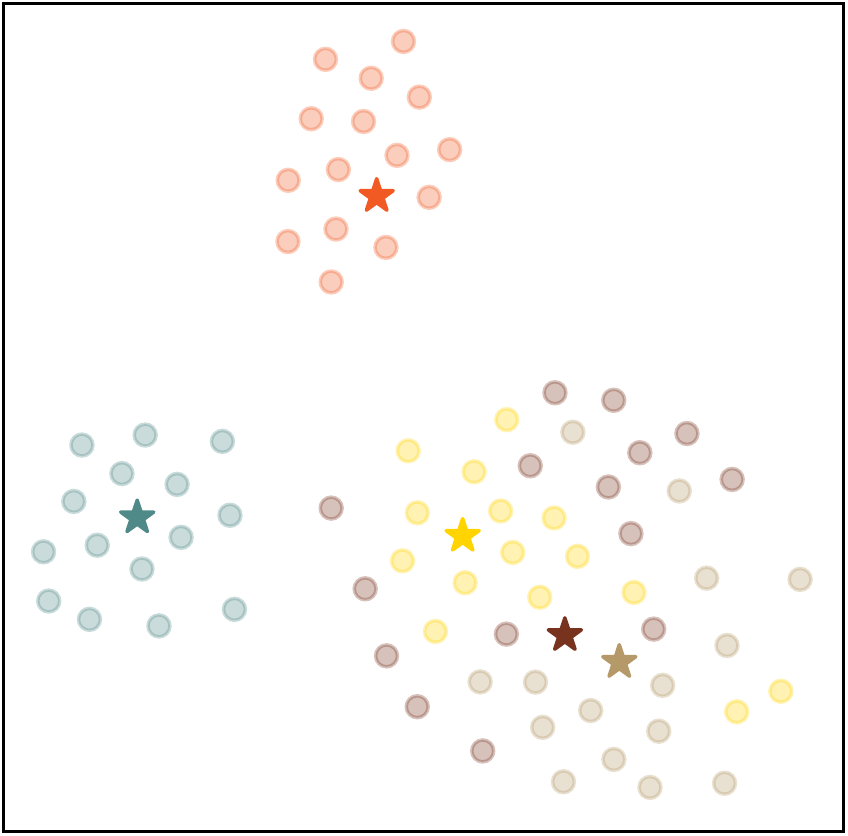}
    \includegraphics[width=0.32\columnwidth]{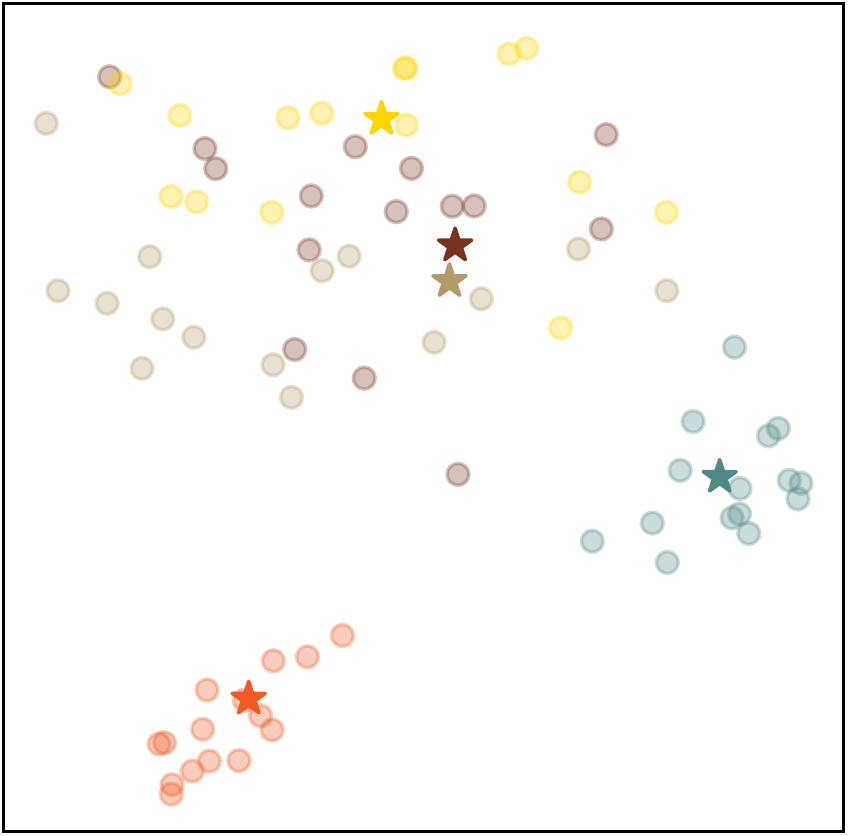}
    \caption{t-SNE visualization results on two tasks.}
    \label{Fig_3}
\end{figure}

We randomly visualize two tasks in Figure~\ref{Fig_3}, from left to right, they represent BML, baseline\emph{-global} and baseline\emph{-local}. Obviously, the prototypes (highlight with \emph{star}) computed by BML are more dispersed in the embedding space, which proves that BML helps to obtain more discriminative features.
\subsection{More analyze of mutual interaction}
In order to further verify the influence of mutual interaction on performance, we design a series of ablation experiments, including the impact of the number of shared blocks and the influence of mutual interaction. Here is the result (S:share, I:independent).
\begin{table}[h]
    \centering
    \caption{Analysis of the number of shared blocks.}
    \label{Tab:si}
    \small
    \renewcommand\tabcolsep{15.0pt}
    \begin{tabular}{p{60pt}p{48pt}p{37pt}}
    \thickhline
    \textbf{Methods}&Accuracy&\emph{Params.}
    \\ \thickhline
    (a) Ensemble & 81.08 $\pm$ 0.31 & 24,930,688
    \\ \hline
    (b) BML$(S^0I^4)$ & 83.10 $\pm$ 0.30 & 24,930,688
    \\ \hline
    (c) BML$(S^1I^3)$ & 83.24 $\pm$ 0.30 & 24,813,504
    \\ \hline
    (d) BML$(S^2I^2)$ & 83.30 $\pm$ 0.29 & 24,249,024
    \\ \hline \rowcolor{gray!20}
    (e) BML$(S^3I^1)$ & \textbf{83.63} $\pm$ \textbf{0.29} & \textbf{21,891,264}
    \\ \thickhline
    \end{tabular}
\end{table}

According to the results shown in Table~\ref{Tab:si}, comparing (a) and (b), the simple integration of baseline-\emph{global} and baseline-\emph{local} without interactive learning has almost no benefit since the difference between two models is relatively large (see in Table 2), while (b) still has good performance, it is mutual interactive learning ensures that the features of the two branches have similarities while maintaining appropriate differences. Comparing (b)-(e), the performance of BML changes relatively gently, which shown the main factor that affects the performance is whether performing binocular mutual learning. To reduce the amount of parameters, BML only separates the last block.

\section{Efficient Implementation of BML}
To fully unleash the power of binocular framework, during training, we adopt uniform sampling strategy. Specifically, a batch contains $N=15$ randomly sampled classes.